\documentclass[letterpaper]{article} 
\usepackage[preprint]{aaai2027}
\usepackage[hyphens]{url}  
\usepackage{graphicx} 
\usepackage{natbib}  
\usepackage{caption} 
\usepackage{algorithm}
\usepackage{algorithmic}
\usepackage{subcaption}
\usepackage{booktabs}
\usepackage{subcaption}
\usepackage[table]{xcolor}
\definecolor{lightcyan}{RGB}{224,255,255}
\usepackage{amsmath}
\usepackage{amssymb}
\usepackage{multirow}
\usepackage{newfloat}
\usepackage{listings}
\DeclareCaptionStyle{ruled}{labelfont=normalfont,labelsep=colon,strut=off} 
\floatstyle{ruled}
\newfloat{listing}{tb}{lst}{}
\floatname{listing}{Listing}

\usepackage{booktabs}

\title{Data DPO: Direct Preference Optimization for Target Model Data Selection in LLM Post-training}

\author{
    Written by AAAI Press Staff\textsuperscript{\rm 1}\thanks{With help from the AAAI Publications Committee.}\\
    AAAI Style Contributions by Peter Patel Schneider,
    Sunil Issar,\\
    J. Scott Penberthy,
    George Ferguson,
    Hans Guesgen,
    Francisco Cruz\equalcontrib\corresponding,
    Marc Pujol-Gonzalez\equalcontrib\corresponding
}
\affiliations{
    \textsuperscript{\rm 1}Association for the Advancement of Artificial Intelligence\\

    1101 Pennsylvania Ave, NW Suite 300\\
    Washington, DC 20004 USA\\
    proceedings-questions@aaai.org
}

\title{Data-DPO: Direct Preference Optimization for Target Model Data Selection in LLM Post-Training}

\author{
    Peng Sun\textsuperscript{\rm 1},
    Yi Yang\textsuperscript{\rm 1},
    Antong Zhang\textsuperscript{\rm 2},
    Chunxiao Li\textsuperscript{\rm 3},\\
    Yanbo Wang\textsuperscript{\rm 4},
    Dianbo Liu\textsuperscript{\rm 5},
    Xin Chen\textsuperscript{\rm 6},
    Kai Yu\textsuperscript{\rm 7},
    Lu Chen\textsuperscript{\rm 7},
    Tianfan Fu\textsuperscript{\rm 1}\corresponding
}

\affiliations{
    \textsuperscript{\rm 1}Nanjing University
    \quad
    \textsuperscript{\rm 2}Brown University
    \quad
    \textsuperscript{\rm 3}Fudan University
    \quad
    \textsuperscript{\rm 4}North University of China
    \\
    \textsuperscript{\rm 5}National University of Singapore
    \quad
    \textsuperscript{\rm 6}Suzhou Laboratory
    \quad
    \textsuperscript{\rm 7}Shanghai Jiao Tong University
}

\begin{document}

\maketitle

\begin{abstract}
Data selection in supervised fine-tuning aims to select a small set of effective samples from large-scale candidate data, reducing training cost while preserving model performance. 
However, existing methods usually treat data value as a relatively static property, and pay limited attention to the compatibility between data and the capability distribution of the target model. 
To address this issue, we propose Data-DPO, a target model-oriented SFT data selection method. Data-DPO observes the local training feedback of the target model on different samples through one-step probing, transforms activation differences among samples into pairwise data preferences, and trains a lightweight reward model to learn target-model-aware data preferences. In the final selection stage, Data-DPO further combines target model preference, external quality scores, and marginal diversity to construct a more stable and effective training subset. Experimental results on Vision-Flan and LLaVA-CoT show that Data-DPO consistently outperforms existing data selection baselines under multiple data budgets and stably surpasses full data training performance.
\end{abstract}


\section{Introduction}

Supervised fine-tuning (SFT) has become a critical step in improving the capabilities of large models~\cite{liu2024improved,lambert2025tulu3pushingfrontiers}. However, as the scale of instruction data continues to grow, directly fine-tuning on the full dataset is not only computationally expensive, but may also fail to yield optimal performance~\cite{NEURIPS2025_0d77ccb5}. Therefore, how to select a small yet effective training subset from large-scale candidate data, so that the model can achieve comparable or even better performance with substantially lower training cost, has become an important problem in SFT~\cite{NEURIPS2025_a59ff5f7}.

Existing data selection methods usually evaluate data value from two perspectives: sample importance and data diversity. The former focuses on the contribution of a sample to model updates or downstream performance~\cite{liu2025less,he2025learning}, while the latter emphasizes distributional coverage and redundancy reduction~\cite{deb2025fishersft,bi2025prism}. Some methods further integrate quality, importance, and diversity to construct a more balanced training subset~\cite{lee2024concept,yan2025coido,yu2025mastering}. Although these methods have achieved promising results across different tasks, most of them still regard data value as a relatively static property, where whether a sample is high-quality is usually determined in advance by the sample itself and the structure of the representation space.

However, in the SFT setting, the value of data should not be determined solely by the intrinsic quality of the sample~\cite{NEURIPS2025_cea04322}. SFT typically does not require a model to learn knowledge from scratch; instead, it further activates, calibrates, and aligns the model's existing capability distribution~\cite{goyal2024context}. Since models differ in architecture, parameter scale, and capability distribution, the same sample may play substantially different roles when used to fine-tune different target models~\cite{hu2025tinyalign}. Some samples may be too difficult for a weaker model, while being exactly suitable for a stronger model. Similarly, an answer that appears to be high-quality may not necessarily match the current learning needs of a specific model~\cite{liu2026data}. Therefore, SFT data selection should not only answer the question, ``Which data is high-quality?'', but should further address, ``Which data is more suitable for the target model?''.

Based on this observation, we propose Data-DPO, a target-model-aware data selection method for SFT. Data-DPO leverages feedback from the target model during a short training process to characterize the compatibility between samples and the current training needs of the model. Specifically, we first select a small set of representative samples from the candidate data, observe the target model's training feedback before and after a one-step update, and transform such feedback into pairwise preference relations between data samples. We then train a lightweight reward model to learn data preferences conditioned on the target model. To prevent the selected subset from being overly biased toward locally easy-to-activate samples, Data-DPO further combines target-model preference, external quality assessment, and data diversity during the final subset construction stage. In this way, it incorporates the target model's own feedback into a strong quality-diversity selection framework.

We evaluate Data-DPO on two datasets, Vision-Flan and LLaVA-CoT, which cover general instruction tuning and reasoning-oriented fine-tuning scenarios, respectively. Experiments are conducted under three data budgets: 5\%, 10\%, and 15\%. We compare Data-DPO with representative baselines, including random selection, importance estimation, diversity-based selection, and hybrid selection methods. The results show that Data-DPO outperforms existing baselines under all data budgets on Vision-Flan, achieving average relative performance of 100.76\%, 102.63\%, and 102.70\% compared with full-data training, respectively. On LLaVA-CoT, Data-DPO achieves 102.73\% and 103.93\% of full-data training performance under the 5\% and 10\% budgets, respectively, substantially outperforming all baselines; under the 15\% budget, it still maintains performance above full-data training.

Our main contributions are as follows:
\begin{itemize}
\item We revisit the problem of SFT data selection from a target-model-conditioned perspective, emphasizing that data value depends not only on the static quality of samples, but also on their compatibility with the capability distribution of the target model.
\item We propose Data-DPO, which constructs data preferences from the target model's own feedback during a short training process, and combines preference, quality, and diversity signals to select a subset better suited to the model.
\item We conduct systematic experiments on Vision-Flan and LLaVA-CoT. The results demonstrate that Data-DPO consistently outperforms representative baselines under multiple data budgets and stably surpasses full-data training performance. Further analyses show that our method is robust across different target models, quality scoring sources, and embedding sources.
\end{itemize}

\section{Related Work}
Existing data selection methods typically select training subsets from two complementary perspectives: sample importance and data diversity. The former estimates the contribution of each training instance to the target task or model update. For example, LESS~\cite{xia2024less}, TIVE~\cite{liu2025less}, ICONS~\cite{wu2024icons}, and OPUS~\cite{wang2026opus} select data based on the gradient similarity between training and validation samples; ScalSelect~\cite{wu2026scalselect} evaluates sample importance through the internal attention behavior of the target model, especially its attention distribution; 
EL2N~\cite{paul2021deep} trains a proxy model on a small subset and identifies informative samples based on their losses under the proxy model. The latter line emphasizes coverage and redundancy reduction. PRISM~\cite{bi2025prism}, FisherSFT~\cite{deb2025fishersft}, SemDeDup~\cite{abbas2023semdedup}, Self-Sup~\cite{sorscher2022beyond}, and D2 Pruning~\cite{maharana2023d2} ensure diversity by embedding-based deduplication or dispersion, while ICONS~\cite{wu2024icons} and INSTAG~\cite{lu2023instag} construct representative subsets by maximizing the coverage of downstream task types. Several studies further combine these two perspectives. CoIDO~\cite{yan2025coido} trains a unified scorer to jointly model data quality and diversity; DataTailor~\cite{yu2025mastering} designs selection criteria from multiple dimensions, including sample informativeness, intra- and inter-cluster relations, and dialogue turns; and COINCIDE~\cite{lee2024concept} extracts multi-layer representations from the target model and samples data from both intra- and inter-cluster structures. Unlike these methods, we do not treat sample utility as static quality, external similarity, or representation-space coverage. Instead, we directly characterize the activation strength of each sample in the target model and transform it into pairwise data preferences for data selection oriented to the target model.

\section{Method}

\subsection{Problem Formulation}
Given an original training set $\mathcal{D}$, data selection aims to select a subset $\mathcal{D}' \subseteq \mathcal{D}$ under a fixed budget $K$, such that the model trained on $\mathcal{D}'$ achieves performance close to, or even better than, training on the full dataset. Let $f(\cdot)$ denote the target-task performance of the model trained on a given dataset. The data selection problem can be formulated as:
\begin{equation}
\max_{\mathcal{D}' \subseteq \mathcal{D}} f(\mathcal{D}')
\quad \mathrm{s.t.} \quad
|\mathcal{D}'| = K .
\end{equation}
Here, $K$ denotes the data budget. This formulation captures the core objective of data selection: identifying a limited subset that maximizes downstream model performance.

\subsection{Data-DPO}
\label{subsec:detail_dpo}
\begin{figure*}[t]
  \centering
  \includegraphics[width=1.0\linewidth]{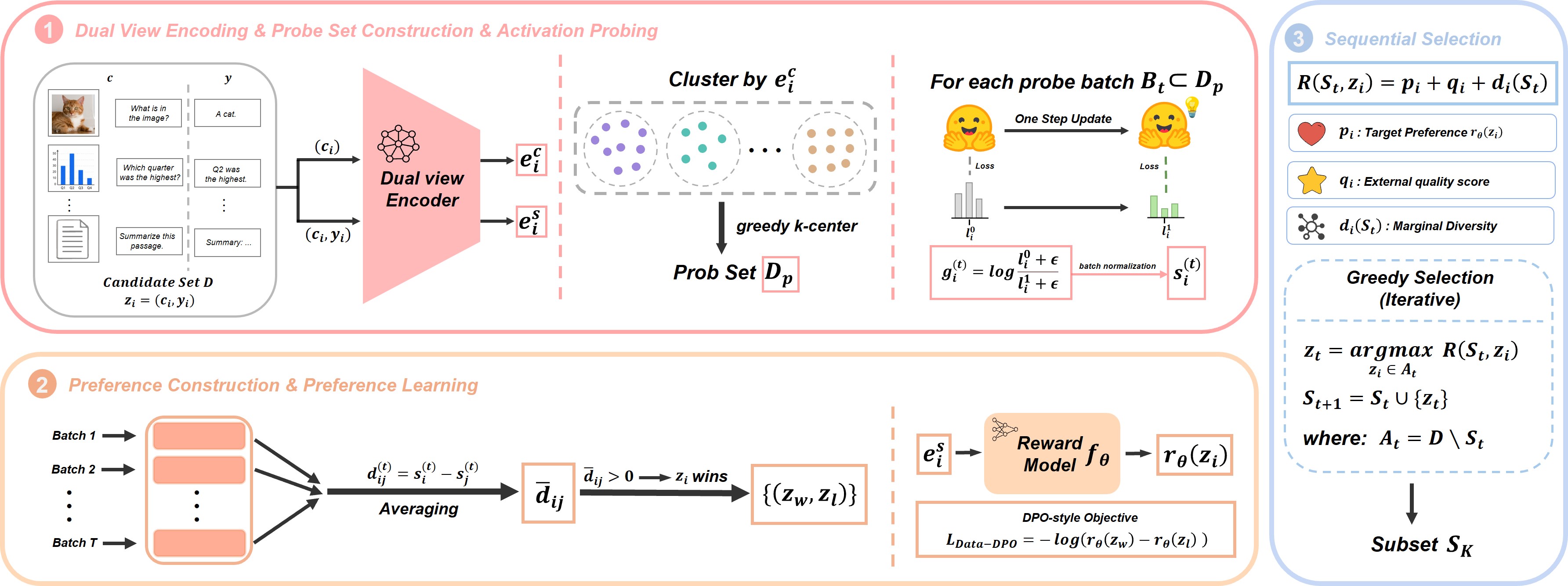}
  \caption{Overview of Data-DPO. Data-DPO constructs a probe set through dual view encoding, derives sample preferences from one-step training feedback of the target model, and learns a target-model-aware reward model. The final subset is then greedily selected by jointly considering preference scores, external quality scores, and marginal diversity.}
  \label{fig:dpo_pipeline}
\end{figure*}

We propose Data-DPO, an SFT data selection method tailored to the target model. Data-DPO constructs pairwise preferences from one-step target model update, trains a reward model from these preferences, and selects the final subset by combining target model preference, general quality, and marginal diversity. The overall pipeline is shown in Figure~\ref{fig:dpo_pipeline}, with implementation details provided in Appendix~\ref{app:data_dpo_implementation}.
\paragraph{Dual View Encoding.}
Given a candidate sample $z_i$, we denote it as a pair: 
\begin{equation}
z_i = (c_i, y_i),
\end{equation}
where $c_i$ is the input condition and $y_i$ is the target response. Data-DPO constructs two complementary representations for each sample. The first is a condition embedding:
\begin{equation}
e_i^c = E_{\mathrm{emb}}(c_i),
\end{equation}
which encodes only the input condition. This representation is used for probe set construction and for measuring marginal diversity during the final selection stage. The second is a supervision embedding:
\begin{equation}
e_i^s = E_{\mathrm{emb}}(c_i, y_i),
\end{equation}
which encodes the full supervised signal of the sample and is used for preference learning and final preference scoring. Both representations are extracted with a frozen embedding model, followed by independent centering and $L_2$-normalization. In the following sections, $e_i^c$ and $e_i^s$ denote the preprocessed embeddings.

\paragraph{Probe Set Construction.}
To construct preference supervision, we first sample a prob subset $\mathcal{D}_p$ from the full candidate set $\mathcal{D}$. Specifically, we cluster $\mathcal{D}$ into $m$ clusters according to the condition embeddings $e_i^c$:
\begin{equation}
\mathcal{D} = \bigcup_{k=1}^{m} \mathcal{C}_k .
\end{equation}

For each cluster $\mathcal{C}_k$, we select $n_k = \lceil \rho |\mathcal{C}_k| \rceil$ samples:
\begin{equation}
\mathcal{D}_p = \bigcup_{k=1}^{m} \mathcal{D}_p^{(k)}, 
\quad |\mathcal{D}_p^{(k)}| = n_k .
\end{equation}

To prevent the probing samples from being concentrated in high-density regions within each cluster, we adopt a greedy $k$-center strategy. For each cluster $\mathcal{C}_k$, we initialize the selected set $S_k$ with the sample closest to the cluster centroid. Then, at each step, we select the sample that is farthest from the current selected set:
\begin{equation}
z^* =
\arg\max_{z_i \in \mathcal{C}_k \setminus S_k}
\min_{z_j \in S_k}
\left(1 - \cos(e_i^c, e_j^c)\right).
\end{equation}

This procedure encourages $\mathcal{D}_p$ to cover a broader condition space, thereby providing more diverse and stable supervision for subsequent preference construction.

\paragraph{Activation Probing.}
To characterize the local activation effect of each sample on the target model $M_0$, we perform $T$ rounds of one-step probing on the prob subset $\mathcal{D}_p$. In the $t$-th probing round, we sample a batch:
\begin{equation}
B_t = \{z_1, \ldots, z_b\}.
\end{equation}

For each sample $z_i \in B_t$, we first compute its SFT loss before the update:
\begin{equation}
\ell_i^0 = \ell(M_0, z_i).
\end{equation}

Starting from the same initial model $M_0$, we then perform one temporary SFT update on $B_t$:
\begin{equation}
M_t' = \mathrm{OneStepUpdate}(M_0, B_t),
\end{equation}
and compute the loss after the update:
\begin{equation}
\ell_i^1 = \ell(M_t', z_i).
\end{equation}

After each probing round, $M_t'$ is discarded, and the next round still starts from $M_0$. Therefore, this procedure captures the training dynamics around the target checkpoint rather than the preference of a continuously trained surrogate model. We define the one-step activation gain of $z_i$ as:
\begin{equation}
g_i^{(t)} =
\log
\frac{\ell_i^0 + \epsilon}
{\ell_i^1 + \epsilon}.
\end{equation}

A larger $g_i^{(t)}$ indicates that the sample yields a greater loss reduction after one update step, suggesting a stronger local activation effect on the current target model. To remove scale differences across probing batches, we further apply batch-level normalization:
\begin{equation}
s_i^{(t)} =
\frac{g_i^{(t)} - \mu_t}
{\sigma_t + \epsilon},
\end{equation}
where $\mu_t$ and $\sigma_t$ denote the mean and standard deviation of activation gains within the current batch. The normalized score $s_i^{(t)}$ measures the activation strength of $z_i$ relative to other samples in the same batch.

\paragraph{Preference Construction.}
Given a probing batch $B_t$, we construct pairwise preferences using the batch-normalized activation scores. For any two samples $z_i, z_j \in B_t$, their relative activation margin is:
\begin{equation}
d_{ij}^{(t)} = s_i^{(t)} - s_j^{(t)} .
\end{equation}

A positive $d_{ij}^{(t)}$ indicates that $z_i$ activates the target model more strongly than $z_j$, yielding $z_i \succ z_j$; otherwise, we set $z_j \succ z_i$. If a pair is compared multiple times, we aggregate its signed margins:
\begin{equation}
\bar{d}_{ij} =
\frac{1}{n_{ij}}
\sum_{t=1}^{n_{ij}} d_{ij}^{(t)} ,
\end{equation}
Here, $n_{ij}$ is the number of valid comparisons. The final preference direction is determined by the sign of $\bar{d}_{ij}$, producing:
\begin{equation}
\mathcal{P} = \{(z_w, z_l)\},
\end{equation}
with $z_w$ and $z_l$ denoting the preferred and dispreferred samples, respectively. This aggregation mitigates batch-context noise and yields more stable preference labels.

\paragraph{Preference Learning.}
Given the preference set $\mathcal{P}$, we train a target-conditioned reward model for data selection. Specifically, we use the supervision embedding $e_i^s$ of each sample $z_i$ as input to a learnable reward model $f_\theta$, implemented as a lightweight residual MLP:
\begin{equation}
r_\theta(z_i) = f_\theta(e_i^s).
\end{equation}
The reward logit induces a data selection policy over the prob subset $\mathcal{D}_p$:
\begin{equation}
\pi_\theta(z_i) =
\frac{\exp(r_\theta(z_i))}
{\sum_{z_j \in \mathcal{D}_p} \exp(r_\theta(z_j))}.
\end{equation}

We optimize $f_\theta$ with a DPO-style objective in the data space. For each preference pair $(z_w,z_l) \in \mathcal{P}$, the objective compares the policy log-ratio against a reference log-ratio:
\begin{equation}
\mathcal{L}_{\mathrm{DPO}}
=
-\mathbb{E}_{(z_w,z_l)\sim\mathcal{P}}
\log \sigma
\left(
\log \frac{\pi_\theta(z_w)}{\pi_\theta(z_l)}
-
\log \frac{\pi_{\mathrm{ref}}(z_w)}{\pi_{\mathrm{ref}}(z_l)}
\right).
\end{equation}
Under the softmax parameterization above, the normalization term is shared by $z_w$ and $z_l$ and cancels out:
\begin{equation}
\log \frac{\pi_\theta(z_w)}{\pi_\theta(z_l)}
=
r_\theta(z_w)-r_\theta(z_l).
\end{equation}

Unlike standard DPO~\cite{rafailov2024directpreferenceoptimizationlanguage}, where the reference policy is induced by a base SFT model, data selection has no natural model-induced reference policy because the candidate set is predefined rather than sampled from a model policy. A non-uniform reference based on quality scores, cluster sizes, or sampling frequencies would inject additional data priors into the objective. Therefore, we use a uniform empirical reference over $\mathcal{D}_p$:
\begin{equation}
\pi_{\mathrm{ref}}(z_i)
=
\frac{1}{|\mathcal{D}_p|},
\quad z_i \in \mathcal{D}_p .
\end{equation}
This gives:
\begin{equation}
\log \frac{\pi_{\mathrm{ref}}(z_w)}
{\pi_{\mathrm{ref}}(z_l)}
=
0.
\end{equation}
Thus, the data-space DPO objective reduces to:
\begin{equation}
\mathcal{L}_{\mathrm{Data\text{-}DPO}}
=
-\mathbb{E}_{(z_w,z_l)\sim\mathcal{P}}
\log \sigma
\left(
r_\theta(z_w)-r_\theta(z_l)
\right).
\end{equation}
After training, $r_\theta(z_i)$ is applied to the full candidate set $\mathcal{D}$ to produce target-conditioned preference scores for all samples.

\paragraph{Sequential Selection.}
After training, we score each sample $z_i \in \mathcal{D}$ with the learned reward model:
\begin{equation}
r_i = r_\theta(z_i).
\end{equation}
Since DPO only constrains pairwise logit differences, we normalize the logits over $\mathcal{D}$ and obtain a bounded preference score:
\begin{equation}
\hat{r}_i =
\frac{r_i - \mu_{\mathcal{D}}}
{\sigma_{\mathcal{D}} + \epsilon},
\quad
p_i =
\sigma\left(\hat{r}_i\right).
\end{equation}

We then select the final subset sequentially by combining target model preference, external quality, and marginal diversity. Given the selected set $S_t$, the diversity gain of a candidate $z_i$ is:
\begin{equation}
d_i(S_t) =
1 -
\max_{z_j \in S_t}
\max\left(0, \cos(e_i^c, e_j^c)\right),
\end{equation}
where $d_i(S_t)=1$ if $S_t=\emptyset$. The step reward is:
\begin{equation} 
R(S_t,z_i)
=
p_i + q_i + d_i(S_t),
\end{equation}
where $q_i \in [0,1]$ is an external quality score. At each step, we select:
\begin{equation}
z_t =
\arg\max_{z_i \in \mathcal{A}_t}
R(S_t,z_i),
\quad
S_{t+1}=S_t\cup\{z_t\},
\end{equation}
where $\mathcal{A}_t$ is the current candidate pool. We repeat this process until $|S_K|=K$, yielding the selected subset $S_K$.
\section{Experiments}
\subsection{Experimental Setup}
\paragraph{Datasets.}
We evaluate Data-DPO on two datasets, Vision-Flan~\cite{xu2024vision} and LLaVA-CoT~\cite{xu2025llava}.
Vision-Flan represents general instruction scenarios, while LLaVA-CoT represents reasoning scenarios.
Together, they allow us to assess the effectiveness of Data-DPO across different task complexities and data distributions.
Detailed dataset descriptions and preprocessing procedures are provided in Appendix~\ref{app:datasets}.

\begin{figure*}[t]
  \centering
  \includegraphics[width=0.8\linewidth]{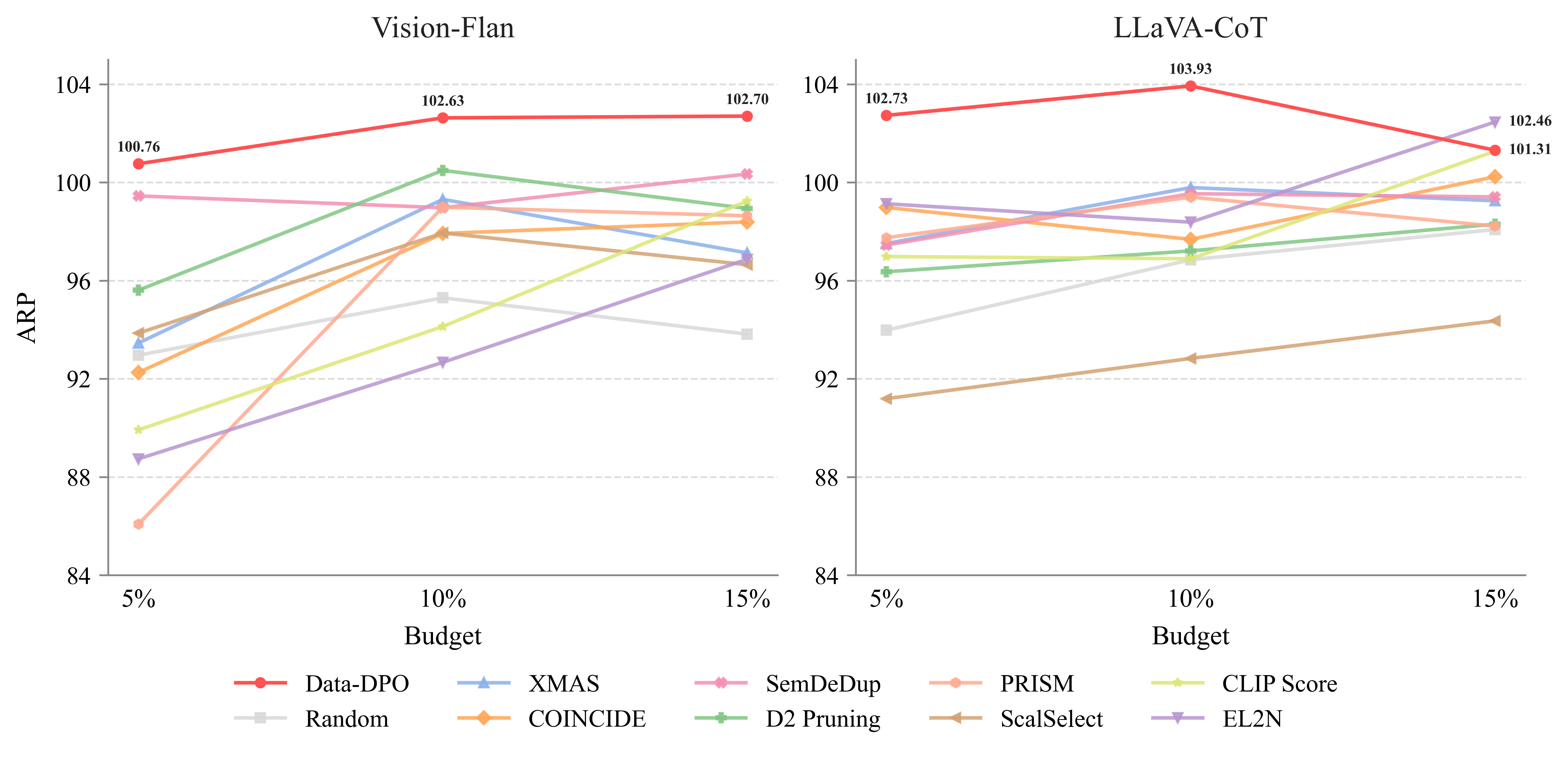}
  \caption{Main results on Vision-Flan and LLaVA-CoT under different data budgets. We report the average relative performance (ARP) of Data-DPO and representative data selection baselines under 5\%, 10\%, and 15\% budgets. Data-DPO achieves consistently strong performance across budgets, outperforming existing baselines in most settings and surpassing full-data training in all budgets on both datasets.}
  \label{fig:main_results}
\end{figure*}

\paragraph{Target Models.}
We use LLaVA-V1.5-7B~\cite{liu2024improved} as the target model on Vision-Flan and Llama-3.2-11B-Vision-Instruct~\cite{grattafiori2024llama} on LLaVA-CoT.
Detailed training hyperparameters are provided in Appendix~\ref{app:training_hyperparameters}.

\paragraph{Data Budgets.}
We evaluate Data-DPO under three data budgets, 5\%, 10\%, and 15\%, for both datasets.

\paragraph{Baselines.}
We compare Data-DPO with Random Selection, XMAS~\cite{naharas2025data}, COINCIDE~\cite{lee2024concept}, SemDeDup~\cite{abbas2023semdedup}, D2 Pruning~\cite{maharana2023d2}, PRISM~\cite{bi2025prism}, ScalSelect~\cite{wu2026scalselect}, CLIP Score~\cite{hessel2021clipscore}, and EL2N~\cite{paul2021deep}.
These baselines cover representative data selection paradigms based on importance estimation, diversity distribution, and the combination of both perspectives.

\paragraph{Quality and Embedding Models.}
To reduce computational overhead, we use lightweight models with vLLM~\cite{kwon2023efficient} for batch scoring.
For quality scoring, we use Qwen3-VL-4B-Instruct~\cite{bai2025qwen3} on Vision-Flan and Qwen3.5-9B~\cite{team2026qwen3} on LLaVA-CoT.
For embedding extraction, we use the lightweight Qwen3-VL-Embedding-2B~\cite{li2026qwen3} model for both datasets.

\paragraph{Evaluation.}
Following the design purpose of each dataset, we evaluate the trained target models on different benchmarks.
Vision-Flan and LLaVA-CoT are each evaluated on a separate set of 12 benchmarks, with the former focusing on general capability evaluation and the latter covering both reasoning and general capability evaluation.
The specific benchmarks used for each dataset and the corresponding detailed results are provided in Appendix~\ref{app:Evaluation}.
To normalize evaluation scales across benchmarks and datasets, we report Average Relative Performance (ARP):
\begin{equation}
\mathrm{ARP} =
\frac{\text{Subset Data Performance}}{\text{Full Data Performance}}
\times 100.
\end{equation}

\subsection{Main Results}

Figure~\ref{fig:main_results} presents the results of Data-DPO on the Vision-Flan and LLaVA-CoT datasets.

On Vision-Flan, Data-DPO achieves ARP of 100.76, 102.63, and 102.70 under the 5\%, 10\%, and 15\% budgets, respectively. These results all surpass the performance of full-data training and are the best among all methods. Compared with the strongest method under each budget, Data-DPO improves ARP by 1.31, 2.14, and 2.35, respectively. Compared with random selection, the improvements are 7.80, 7.33, and 8.88 ARP, respectively.

On LLaVA-CoT, Data-DPO also demonstrates clear advantages. Under the 5\% and 10\% budgets, Data-DPO obtains ARP of 102.73 and 103.93, respectively, significantly outperforming all methods. Compared with the strongest method under the corresponding budgets, Data-DPO improves ARP by 3.60 and 4.14, respectively. When the budget increases to 15\%, EL2N achieves an ARP of 102.46, slightly higher than the 101.31 achieved by Data-DPO. Nevertheless, Data-DPO still outperforms all other methods except EL2N and continues to achieve relative performance above full-data training.

Overall, Data-DPO demonstrates strong effectiveness and stability across different data budgets. On both datasets, Data-DPO consistently outperforms full-data training, and it substantially outperforms other methods in the overall comparison. These results jointly validate the core hypothesis of Data-DPO: data selection should not rely solely on static data quality or representation space coverage, but should explicitly model the target model's preferences over data.

\begin{figure}[t]
  \centering
  \includegraphics[width=0.85\linewidth]{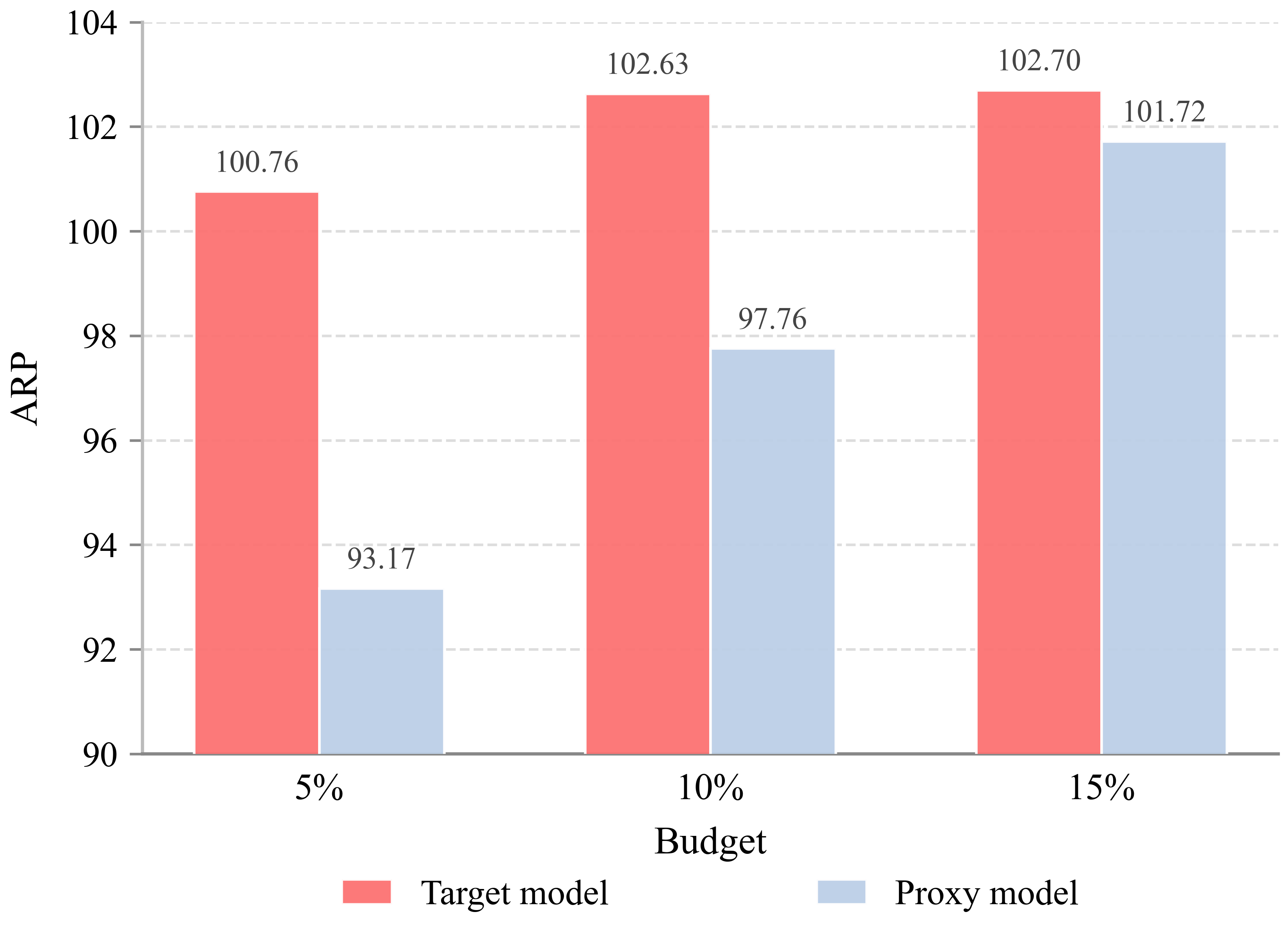}
  \caption{Effect of proxy models in selection. Data-DPO performs consistently better when preferences are constructed from the original target model rather than a 5\% checkpoint, suggesting that proxy models may provide misaligned selection signals.}
  \label{fig:fig_proxy}
\end{figure}

\section{Analysis and Ablation Studies}
\subsection{Analysis of Proxy Models in Selection}

We analyze a key design of Data-DPO: performing one-step probing directly on the original target model. Many existing data selection methods rely on proxy models trained with a small amount of data to provide selection signals~\cite{xia2024less,lee2024concept,naharas2025data}, implicitly assuming that such signals can approximate the data needs of the original target model. However, this assumption may be unreliable in the SFT setting. SFT typically aims to activate the latent capabilities already acquired during pretraining; even a small amount of SFT can shift the model's capability distribution, leading to data preferences that are inconsistent with those of the original target model.

To verify this, we replace the original LLaVA model used for one-step probing in Data-DPO with a checkpoint trained on randomly sampled 5\% Vision-Flan data, while keeping all other components unchanged. As shown in Figure~\ref{fig:fig_proxy}, using this checkpoint to construct preferences reduces ARP from 100.76, 102.63, and 102.70 to 93.17, 97.76, and 101.72 under the 5\%, 10\%, and 15\% budgets, respectively. The drop is especially clear under smaller budgets. This result suggests that selection signals from a proxy model can be affected by distribution shift and may fail to reflect the true data needs of the original target model. Therefore, directly constructing data preferences on the original target model is important for the stable performance of Data-DPO.

\begin{figure}[t]
  \centering
  \includegraphics[width=0.75\linewidth]{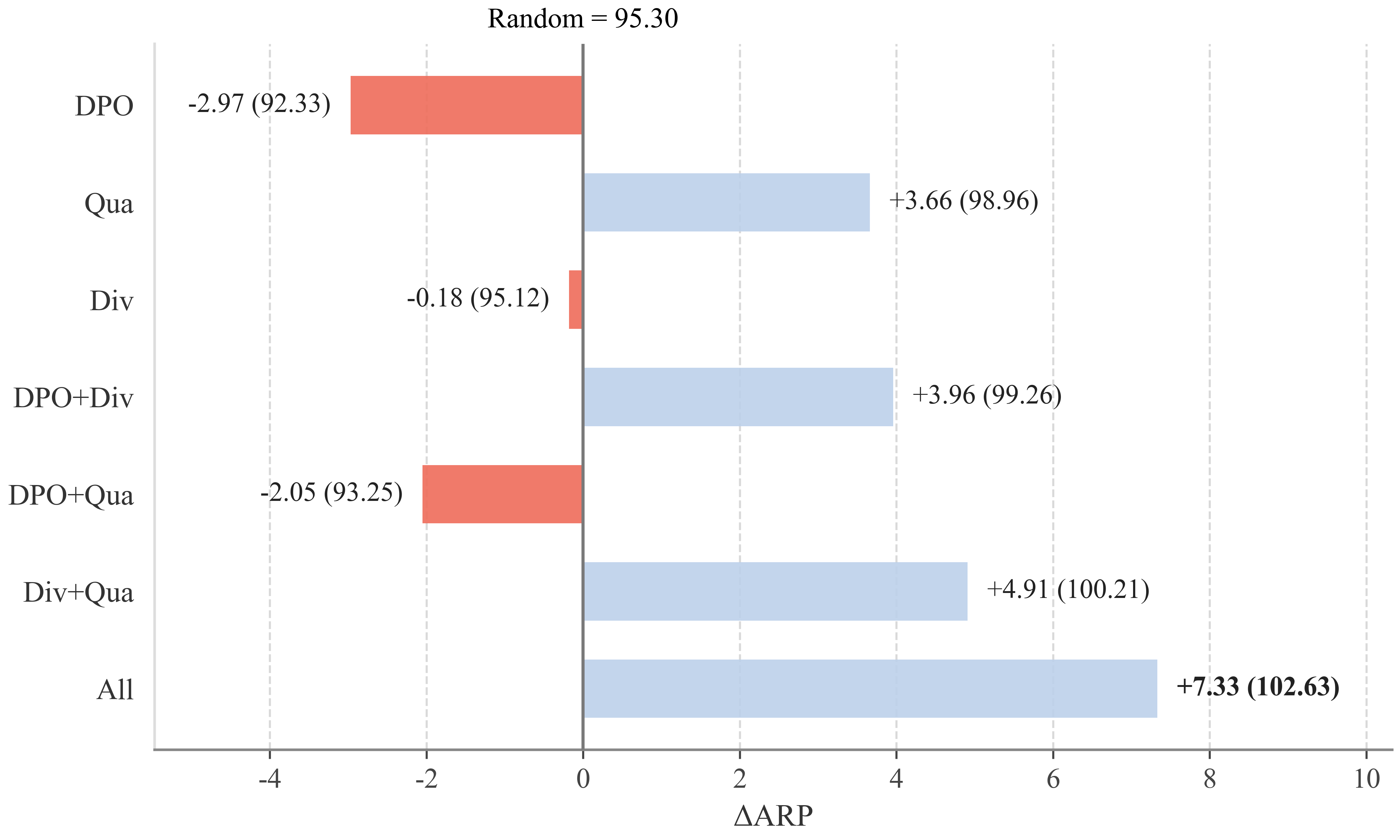}
  \caption{Ablation of the selection reward. We report the ARP improvement over random selection, with the absolute ARP shown in parentheses. Combining target model preference, quality, and marginal diversity achieves the best performance, highlighting the complementarity of three signals.}
  \label{fig:fig_ablation}
\end{figure}

\subsection{Ablation on Selection Reward}

We analyze the role of the three signals in the final selection reward, including the target model preference learned by Data-DPO, the quality score, and marginal diversity. 

Figure~\ref{fig:fig_ablation} shows the ablation results on Vision-Flan under the 10\% budget. When using only the target model preference, the ARP is only 92.33, lower than random selection with 95.30. This is because the preference score tends to select samples that can quickly activate the model. Without quality and distribution constraints, the selected data may concentrate in a narrow and easy region, making it difficult to form an effective training distribution. For pairwise combinations, adding marginal diversity improves DPO + Diversity to 99.26, showing that diversity can reduce the concentration caused by preference-based selection. Quality + Diversity reaches 100.21, indicating that the combination of quality and coverage is a stable signal for data selection. In contrast, DPO + Quality only reaches 93.25, suggesting that adding quality constraints alone is still insufficient to address the concentration of preferred samples. Finally, combining all three signals achieves the best ARP of 102.63, clearly outperforming all single signal and pairwise configurations. This demonstrates the complementarity among target model preference, sample quality, and marginal diversity, which is crucial for the stable performance of Data-DPO.

\subsection{Robustness Across Target Models}

We evaluate the applicability of Data-DPO across target models by testing LLaVA-v1.5-13B~\cite{liu2024improved} on Vision-Flan and Qwen2VL-2B-Instruct~\cite{wang2024qwen2} on LLaVA-CoT, covering both variation in parameter scale within the same model family and variation across different model families.

\begin{figure}[t]
  \centering
  \includegraphics[width=1.0\linewidth]{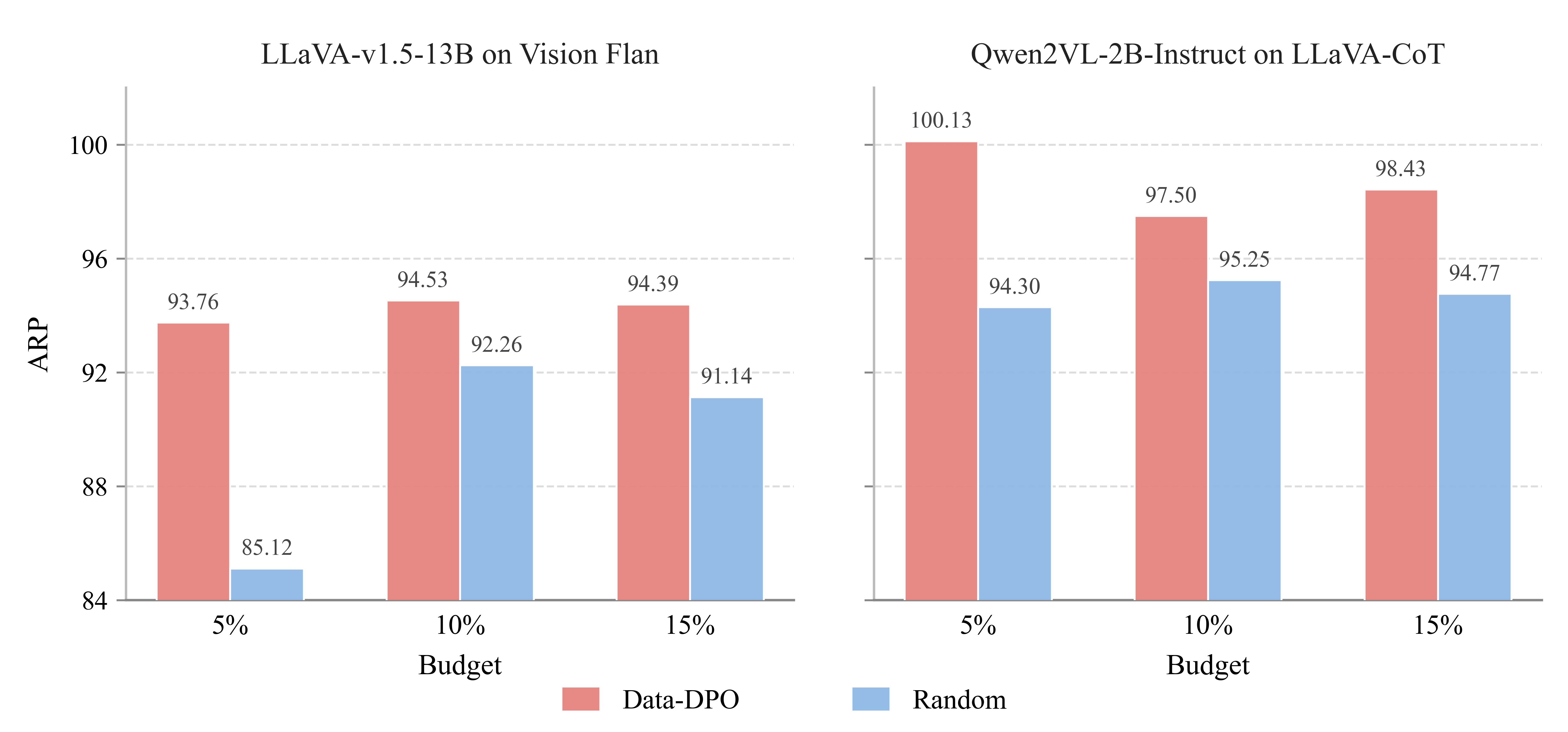}
  \caption{Robustness across target models. We evaluate Data-DPO by replacing the target models with LLaVA-v1.5-13B on Vision-Flan and Qwen2VL-2B-Instruct on LLaVA-CoT, the results show that Data-DPO maintains stable performance across different model scales and model families.}
  \label{fig:fig_model_adaptation}
\end{figure}

As shown in Figure~\ref{fig:fig_model_adaptation}, Data-DPO consistently outperforms random selection in both settings. On Vision-Flan with LLaVA-v1.5-13B, Data-DPO achieves ARP of 93.76, 94.53, and 94.39 under the 5\%, 10\%, and 15\% budgets, improving over random selection by 8.64, 2.27, and 3.25, respectively. On LLaVA-CoT with Qwen2VL-2B-Instruct, Data-DPO also obtains stable gains, reaching 100.13, 97.50, and 98.43 ARP under the three budgets, with improvements of 5.83, 2.25, and 3.66 over random selection. These results show that the target model preferences learned by Data-DPO are not limited to a single model scale or model family, indicating good generalization across different target models.

\subsection{Robustness to Quality Signals}

We analyze the effect of external quality scores as an auxiliary signal in Data-DPO. Since the final selection reward includes a quality term, it is important that the method does not rely on a specific source of quality scores. To examine this, we replace the default quality scoring model with LLaVA-OneVision-1.5-4B-Instruct~\cite{an2025llava} on Vision-Flan, while keeping all other settings unchanged.

As shown in Figure~\ref{fig:quality_signal_robustness}, Data-DPO still clearly outperforms random selection under all budgets when using the alternative quality scores. Under the 5\%, 10\%, and 15\% budgets, it achieves ARP of 101.23, 100.53, and 103.76, improving over random selection by 8.27, 5.23, and 9.94, respectively. This indicates that the quality score mainly serves as an auxiliary constraint in Data-DPO. The final selection performance does not depend on a single quality score source, but comes from the joint effect of target model preference, quality constraint, and marginal diversity.

\subsection{Robustness to Embedding Sources}

We analyze the sensitivity of Data-DPO to the source of embeddings. Data-DPO uses embeddings for probe set construction, reward model input representation, and marginal diversity computation. Therefore, if the method depends heavily on a specific embedding model, its applicability may be limited. To examine this, we replace the embedding model with Qwen3-VL-Embedding-8B~\cite{li2026qwen3} on Vision-Flan, while keeping all other settings unchanged.

\begin{figure}[t]
  \centering

  \begin{subfigure}[t]{0.48\linewidth}
    \centering
    \includegraphics[width=\linewidth]{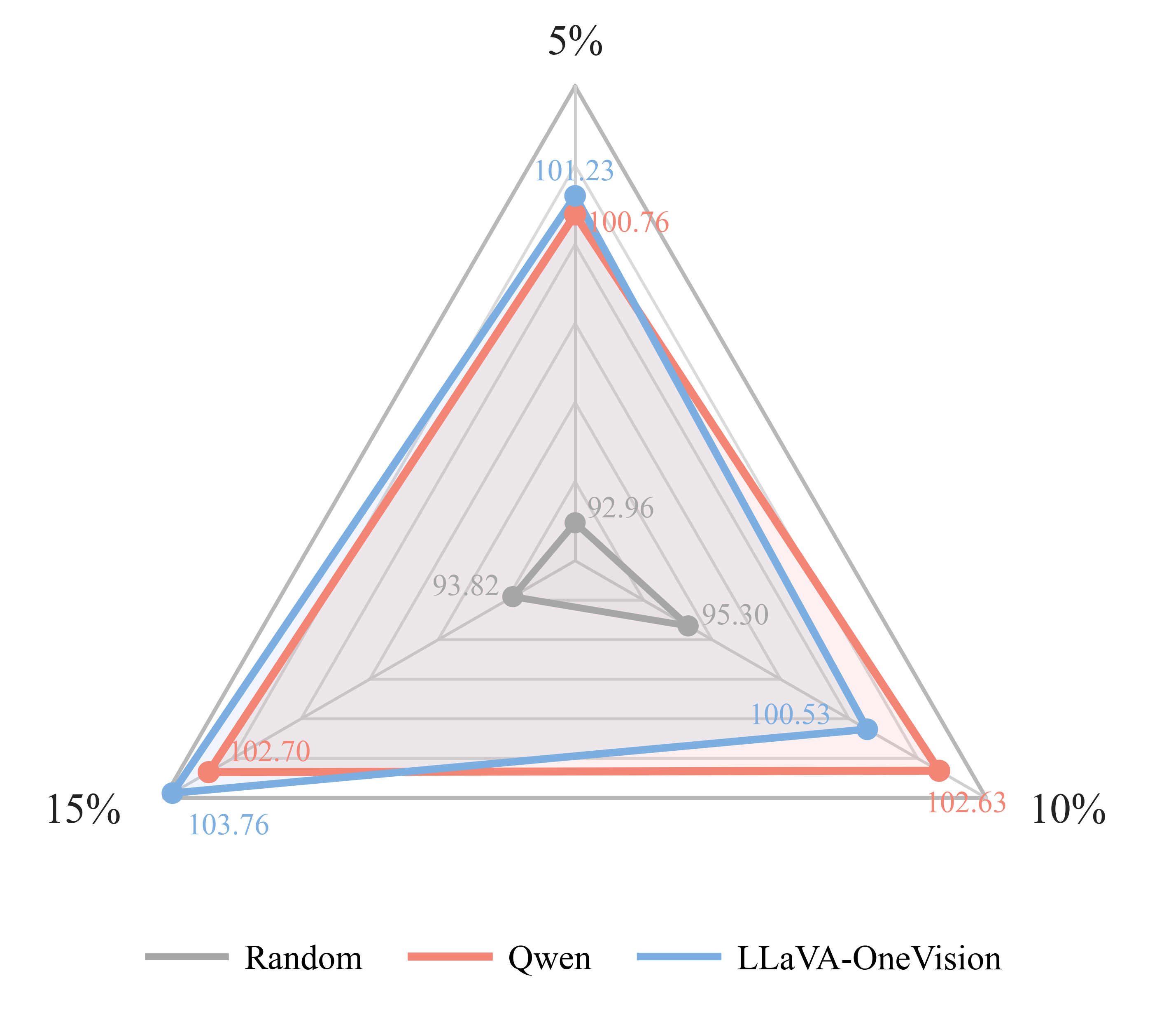}
    \caption{}
    \label{fig:quality_signal_robustness}
  \end{subfigure}
  \hfill
  \begin{subfigure}[t]{0.48\linewidth}
    \centering
    \includegraphics[width=\linewidth]{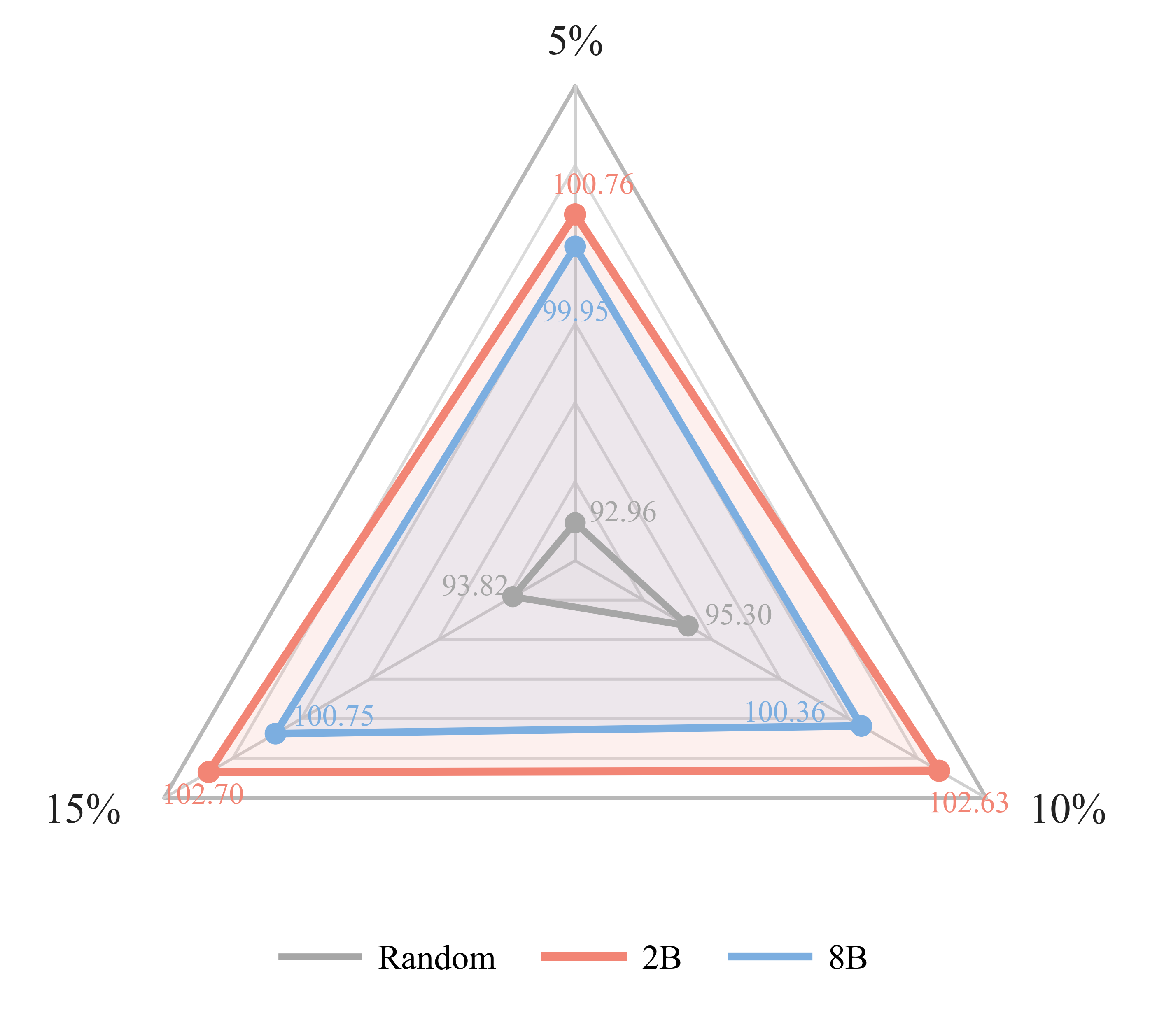}
    \caption{}
    \label{fig:embedding_source_robustness}
  \end{subfigure}

  \caption{Robustness analysis of Data-DPO. (a) Robustness to quality signals. Data-DPO's effectiveness is not tied to a specific source of quality signals. (b) Robustness to embedding sources. Data-DPO's effectiveness is not tied to a specific embedding source.}
  \label{fig:data_dpo_robustness}
\end{figure}

As shown in Figure~\ref{fig:embedding_source_robustness}, Data-DPO still consistently outperforms random selection under all budgets after replacing the embedding model. Under the 5\%, 10\%, and 15\% budgets, it achieves ARP of 99.95, 100.36, and 100.75, improving over random selection by 6.99, 5.06, and 6.93, respectively. Although different embedding models introduce some performance variation, Data-DPO maintains stable gains. This indicates that embeddings mainly serve as auxiliary representation signals in Data-DPO, and the final performance does not depend on a single embedding source. Instead, it is jointly determined by target model preference, quality constraint, and marginal diversity.

\begin{figure}[t]
  \centering
  \includegraphics[width=0.7\linewidth]{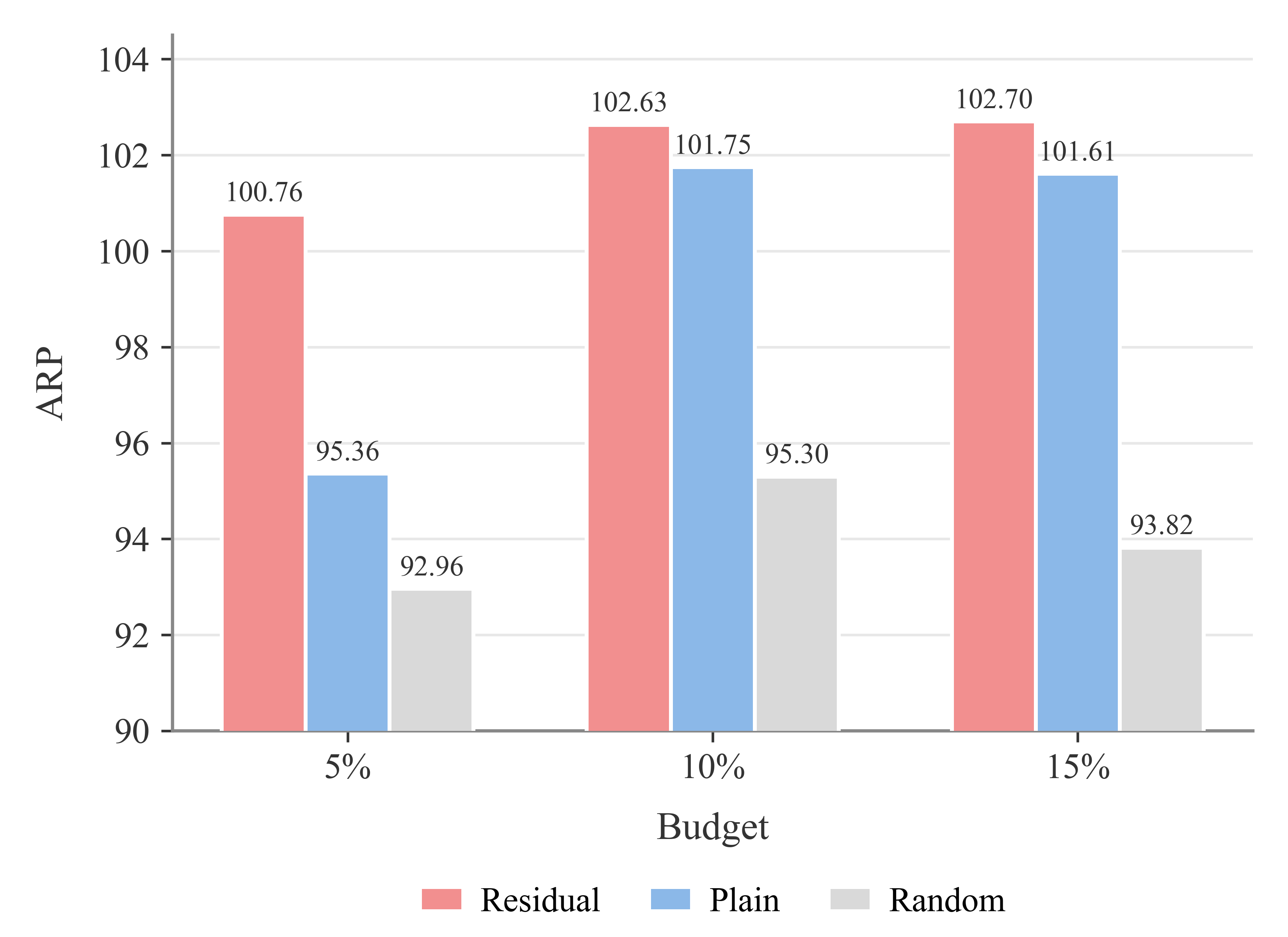}
  \caption{Sensitivity to reward model capacity. We evaluate Data-DPO by replacing the default residual MLP reward model with a simpler plain MLP. The plain MLP still achieves competitive performance, suggesting that the effectiveness of Data-DPO mainly comes from target-model-aware preference supervision rather than reward model capacity alone.}
  \label{fig:fig_reward_model}
\end{figure}

\subsection{Sensitivity to Reward Model Capacity}
\label{app:reward_model}
We analyze the sensitivity of Data-DPO to the reward model capacity by replacing the default residual MLP with a simpler plain MLP, while keeping all other settings unchanged.

Figure~\ref{fig:fig_reward_model} shows the results on Vision-Flan, The plain MLP consistently outperforms random selection across all budgets and surpasses full-data training under the 10\% and 15\% budgets, indicating that Data-DPO does not rely on a complex reward model. Under the 5\% budget, it achieves 95.36 ARP, improving over random selection by 2.40 but lagging behind the residual MLP at 100.76, suggesting that stronger preference fitting is more helpful when the budget is extremely limited. With larger budgets, the gap narrows: the plain MLP reaches 101.75 and 101.61 ARP under the 10\% and 15\% budgets, compared with 102.63 and 102.70 for the residual MLP. Overall, the plain MLP can already learn effective selection signals from pairwise preferences constructed from target model feedback, while the residual MLP provides higher and more stable performance. This suggests that the gains of Data-DPO mainly come from target-model-aware preference supervision rather than reward model capacity alone.

\section{Conclusion and Limitations}

\paragraph{Conclusion.}
We propose Data-DPO, a target model-oriented SFT data selection method. Unlike existing methods that treat data value as static quality, Data-DPO uses the training feedback of the target model during a one step update to construct data preferences among samples, and learns target-model-aware data selection signals through a lightweight reward model. During final subset construction, Data-DPO further combines target model preference, external quality assessment, and marginal diversity to obtain a more stable and effective data distribution. Experimental results show that Data-DPO achieves substantially better performance than most baselines on Vision-Flan and LLaVA-CoT under small data budgets, and stably surpasses full data training performance in multiple settings.

\paragraph{Limitations.}
Despite its effectiveness, Data-DPO has several limitations. First, its performance may be affected by biases in external quality signals, embedding sources, and the target model’s preferences, leading to potential degradation when these signals are highly biased. Second, the method focuses on quality and preference modeling under distribution preserving settings, which may result in limited effectiveness when the full dataset exhibits substantial distribution shifts. Finally, although it avoids training additional proxy models, it still requires additional computation on the prob set, leaving room for further efficiency improvements.

\clearpage
\appendix
\section{Implementation Details of Data-DPO}
\label{app:data_dpo_implementation}

This section supplements the implementation details and hyperparameters omitted from Section~\ref{subsec:detail_dpo} to ensure experimental reproducibility.

\paragraph{Dual View Encoding.}
For each sample $z_i=(c_i,y_i)$, the condition embedding $e_i^c$ is obtained by feeding only $c_i$ into the frozen embedding model, while the supervision embedding $e_i^s$ is obtained by feeding the concatenated pair $(c_i,y_i)$ into the embedding model.

\paragraph{Probe Set Construction.}
When constructing the prob subset $\mathcal D_p$, we adopt spherical clustering with 2000 clusters for Vision-Flan and 1000 clusters for LLaVA-CoT. Within each cluster, probing samples are selected using a fixed probe ratio of $\rho=0.05$.

\paragraph{Activation Probing.}
We fix the number of probing rounds to $T=16$, and set the probing batch size to $b=1024$ for each round. During activation probing, the training hyperparameters are kept the same as those used in the SFT stage. The detailed SFT training configuration is provided in Appendix~\ref{app:training_hyperparameters}.

\paragraph{Preference Construction.}
Within each probing batch, we enumerate all unordered sample pairs $(i,j)$, where $i<j$.

\paragraph{Preference Learning.}
We use a lightweight residual MLP as the reward model, which takes the supervision embedding $e_i^s$ as input and outputs a scalar reward logit $r_\theta(z_i)$. The input dimension is 2048. We first project the input to a hidden dimension of 1024 with a linear layer $\mathrm{Linear}(2048,1024)$. The projected representation is then passed through 4 residual MLP blocks. Each block consists of $\mathrm{LayerNorm}(1024)$, $\mathrm{Linear}(1024,2048)$, GELU activation\textcolor{red}{reference}, $\mathrm{Dropout}(0.1)$, $\mathrm{Linear}(2048,1024)$, and dropout, followed by a residual connection. The final output is produced by $\mathrm{LayerNorm}(1024)$ and $\mathrm{Linear}(1024,1)$.  The reward model is trained with the AdamW optimizer. The learning rate is set to $1\times 10^{-3}$, and the weight decay is set to $1\times 10^{-4}$. The training batch size is 4096, and the number of training epochs is 5. We split the constructed preference pairs into 95\% for training and 5\% for validation. The training objective is the Data-DPO objective defined in the main text, namely the pairwise preference optimization objective under a uniform empirical reference.

\begin{figure*}[t]
    \centering
    \includegraphics[width=0.5\linewidth]{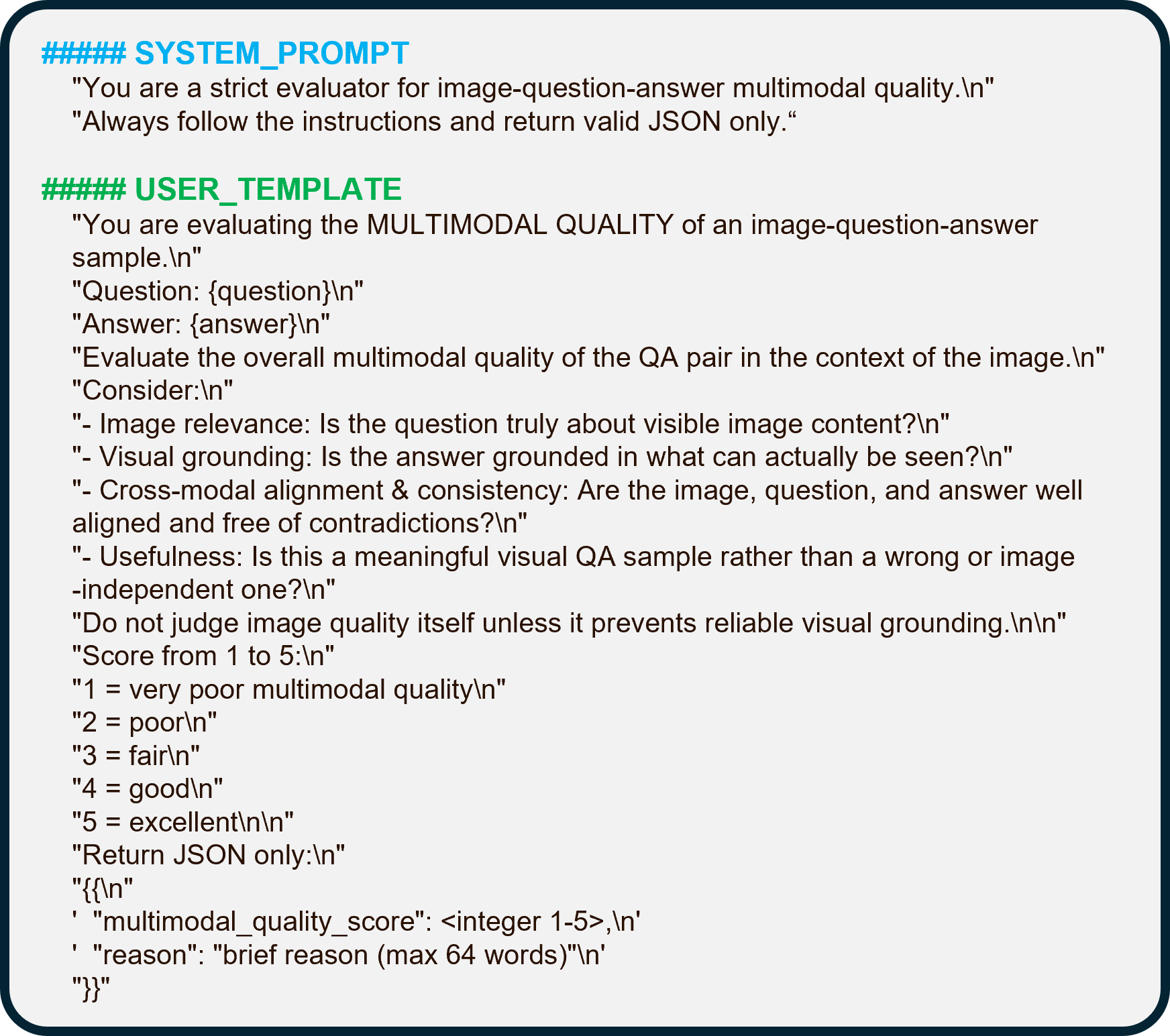}
    \caption{Scoring prompt for estimating the quality score on Vision-Flan.}
    \label{fig:prompt_vf_mm}
\end{figure*}

\begin{figure*}[t]
    \centering
    \includegraphics[width=0.5\linewidth]{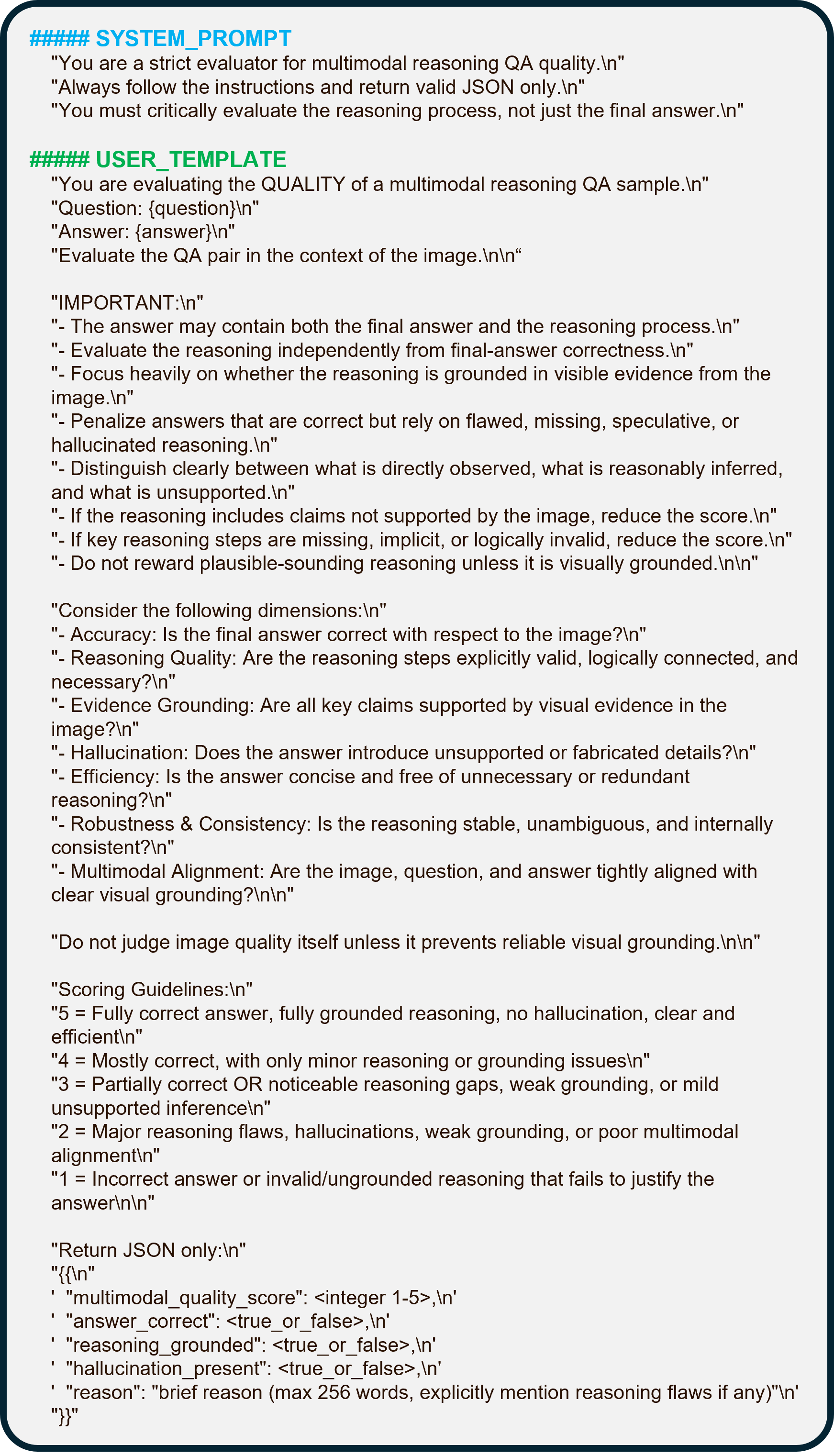}
    \caption{Scoring prompt for estimating the quality score on LLaVA-CoT.}
    \label{fig:prompt_llavacot}
\end{figure*}

\paragraph{Sequential Selection.}
The external quality score is obtained by feeding each sample into a multimodal model, as described in Section~\ref{subsec:detail_dpo}. For each sample, we use a discrete rating scale from 1 to 5, and normalize the rating to $[0,1]$. Specifically, the scores $1,2,3,4,5$ are mapped to $0,0.25,0.5,0.75,1$, respectively. The prompts used for quality scoring are shown in Figure~\ref{fig:prompt_vf_mm} and Figure~\ref{fig:prompt_llavacot}.

\section{Datasets and Preprocessing Details}
\label{app:datasets}
\subsection{Datasets}
\paragraph{Vision-Flan.}
Vision-Flan~\cite{xu2024vision} is a human-curated visual instruction tuning dataset that targets broad task coverage and strong cross-task generalization for vision-language models. The dataset is built upon publicly available academic vision benchmarks and reorganizes them into 187 fine-grained visual tasks with manually written task instructions. Each task contains roughly 1,000 examples, resulting in about 186K samples in total. Compared with instruction data generated purely by synthetic pipelines, Vision-Flan emphasizes expert annotation and validation, which helps reduce instruction artifacts and unreliable supervision.

\paragraph{LLaVA-CoT.}

LLaVA-CoT-100k~\cite{xu2025llava} is an image-text reasoning instruction dataset designed to improve the structured reasoning ability of vision-language models. It integrates approximately 99K image-question-answer pairs from a mixture of general-purpose and science-targeted VQA datasets. Each sample is annotated with a structured reasoning response consisting of summary, caption, reasoning, and conclusion stages.


\subsection{Preprocessing Procedures}
For Vision-Flan, we use the original dataset without additional filtering or format conversion. For LLaVA-CoT, we remove samples without associated images, resulting in 98,572 image-text reasoning samples. The reasoning process and final answer in each response are wrapped with \verb|<think></think>| and \verb|<answer></answer>|, respectively.

\section{Training Hyperparameters}
\label{app:training_hyperparameters}

Table~\ref{tab:training_hyperparameters} summarizes the training hyperparameters used for all target models. 
For LLaVA-V1.5-7B and LLaVA-V1.5-13B, we follow the hyperparameter settings from the official LLaVA training code. For Qwen2-VL-2B-Instruct and Llama-3.2-11B-Vision-Instruct, since both models are already instruction-tuned checkpoints, we use a conservative learning rate of $1\times10^{-5}$ for full-parameter fine-tuning. 
We set the batch size to 64 for these two models because using a batch size of 128 leads to too few optimization steps within one epoch, both for the full dataset and for subset training, making the training insufficient.

\section{Evaluation}
\label{app:Evaluation}

\subsection{Evaluation Benchmarks}
\label{sec:benchmarks}

We choose benchmarks according to the modality and target capability of each dataset. For Vision-Flan, we use benchmarks covering general multimodal understanding, visual question answering, OCR, document understanding, chart understanding, and science-diagram reasoning. For LLaVA-CoT, we combine multimodal mathematical and logical reasoning benchmarks with general vision-language benchmarks to evaluate both reasoning ability and overall multimodal robustness. For Qwen2-VL-2B-Instruct trained on LLaVA-CoT, we observe that full-data training still yields extremely poor performance on We-Math~\cite{qiao2025we}, LogicVista~\cite{xiao2024logicvista}, DynaMath~\cite{zou2024dynamath}, and MMStar~\cite{chen2024we}, making these benchmarks less informative for comparing data selection methods under this target model. Therefore, for this model we evaluate on the remaining eight benchmarks.The full benchmark list is shown in Table~\ref{tab:benchmark_list}.

\subsection{Detailed Evaluation Results}
\label{sec:detailed_results}
For the main experiments, Tables~\ref{tab:full_comparison_vf_dpo} and~\ref{tab:llavacot_llama32_11b_comparison} present the full benchmark-level comparison results on Vision-Flan and LLaVA-CoT, respectively, covering Data-DPO and all data selection baselines. We further provide benchmark-level results for the analysis and ablation studies discussed in the main paper and appendix. Specifically, Table~\ref{tab:proxy_model_5ckpt_ablation} reports the results of the proxy-model analysis. Table~\ref{tab:reward_ablation_vf} presents the ablation results on the selection reward. Table~\ref{tab:robustness_vf_llava13b} shows the Vision-Flan results with LLaVA-V1.5-13B as the target model, while Table~\ref{tab:robustness_cot_qwen2vl2b} shows the LLaVA-CoT results with Qwen2-VL-2B-Instruct. Table~\ref{tab:quality_signal_vf} reports the results under different quality signal sources. Table~\ref{tab:embedding_source_robustness_vf} reports the results under different embedding sources. Finally, Table~\ref{tab:reward_model_robustness_vf} reports the results with a simplified reward model.

\section{Time Cost Analysis}
\label{app:timecost}
We measure the time cost of different data selection methods on LLaVA-CoT. Table~\ref{tab:time_cost} reports the GPU hours required by each method on a NVIDIA A6000. Data-DPO takes approximately 19.0 GPU hours in total. Although its computational cost is slightly higher than other baselines, it remains acceptable compared with full SFT training and achieves the best or near-best performance under most data budgets.

\clearpage

\begin{table*}[t]
\centering
\small
\caption{Training hyperparameters for target models.}
\setlength{\tabcolsep}{3.5pt}
\resizebox{0.95\textwidth}{!}{
\begin{tabular}{lccccccccc}
\toprule
\textbf{Model} & \textbf{Strategy} & \textbf{Epoch} & \textbf{Batch size} & \textbf{LR} & \textbf{Optimizer} & \textbf{LR scheduler} & \textbf{Warmup ratio} & \textbf{LoRA rank} & \textbf{LoRA alpha} \\
\midrule
LLaVA-V1.5-7B & LoRA & 1 & 128 & $2\times10^{-4}$ & AdamW & Cosine & 0.03 & 128 & 256 \\
LLaVA-V1.5-13B & LoRA & 1 & 128 & $2\times10^{-4}$ & AdamW & Cosine & 0.03 & 128 & 256 \\
Qwen2-VL-2B-Instruct & Full & 1 & 64 & $1\times10^{-5}$ & AdamW & Cosine & 0.03 & -- & -- \\
Llama-3.2-11B-Vision-Instruct & Full & 1 & 64 & $1\times10^{-5}$ & AdamW & Cosine & 0.03 & -- & -- \\
\bottomrule
\end{tabular}
}
\label{tab:training_hyperparameters}
\end{table*}

\begin{table*}[t]
\centering
\small
\caption{Evaluation benchmarks used for each dataset.}
\label{tab:eval_benchmarks}
\begin{tabular}{cl}
\toprule
\textbf{Dataset} & \textbf{Benchmark} \\
\midrule
\multirow[c]{12}{*}{Vision-Flan}
& GQA~\cite{hudson2019gqa} \\
& VizWiz~\cite{gurari2018vizwiz} \\
& TextVQA~\cite{singh2019towards} \\
& ScienceQA-IMG~\cite{saikh2022scienceqa} \\
& MME~\cite{liang2024survey} \\
& MMBench~\cite{liu2024mmbench} \\
& AI2D~\cite{kembhavi2016diagram} \\
& ChartQA~\cite{masry2022chartqa} \\
& DocVQA~\cite{mathew2021docvqa} \\
& InfoVQA~\cite{mathew2022infographicvqa} \\
& MMStar~\cite{chen2024we} \\
& OCRBench~\cite{liu2024ocrbench} \\
\midrule

\multirow[c]{12}{*}{LLaVA-CoT}
& MATH-Vision~\cite{wang2024measuring} \\
& We-Math~\cite{qiao2025we} \\
& LogicVista~\cite{xiao2024logicvista} \\
& DynaMath~\cite{zou2024dynamath} \\
& MMStar~\cite{chen2024we} \\
& MME~\cite{liang2024survey} \\
& MMBench-EN~\cite{liu2024mmbench} \\
& ScienceQA-IMG~\cite{saikh2022scienceqa} \\
& AI2D~\cite{kembhavi2016diagram} \\
& ChartQA~\cite{masry2022chartqa} \\
& InfoVQA~\cite{mathew2022infographicvqa} \\
& OCRBench~\cite{liu2024ocrbench} \\

\bottomrule
\end{tabular}
\label{tab:benchmark_list}
\end{table*}

\begin{table*}[htbp]
  \centering
  \caption{Main experimental results on Vision-Flan using LLaVA-v1.5-7B as the target model. The best result in each column is highlighted in bold.}
  \label{tab:full_comparison_vf_dpo}

  \begin{subtable}{\textwidth}
    \centering
    \caption{5\% data subset.}
    \label{tab:vf_dpo_subset_5}
    \resizebox{\textwidth}{!}{%
      \begin{tabular}{lcccccccccccccc}
      \toprule
      \textbf{Method} & \textbf{GQA} & \textbf{VizWiz} & \textbf{TextVQA} & \textbf{SQA-I} & \textbf{MME} & \textbf{MMB-CN} & \textbf{MMB-EN} & \textbf{AI2D} & \textbf{ChartQA} & \textbf{DocVQA} & \textbf{InfoVQA} & \textbf{MMStar} & \textbf{OCRBench} & \textbf{ARP} \\
      \midrule
        Full Data & 47.30 & 55.88 & 35.83 & 61.43 & 1270.40 & 50.26 & 55.67 & 52.14 & 15.68 & 15.80 & 15.53 & 35.32 & 26.20 & 100.00 \\
        \midrule
        Random & 42.93 & 55.79 & 36.54 & 60.34 & 1098.02 & 36.40 & 26.19 & 37.82 & 15.96 & 16.46 & 19.99 & 34.77 & 27.80 & 92.96 \\
        XMAS & 40.38 & 54.15 & 37.06 & 62.42 & 705.48 & 39.95 & 39.95 & 41.22 & 15.40 & 17.98 & 19.80 & 34.37 & 27.50 & 93.46 \\
        COINCIDE & 42.15 & 53.48 & 38.00 & 55.03 & 1000.71 & 29.12 & 41.24 & 36.40 & 16.64 & 17.89 & 17.79 & 33.51 & 28.70 & 92.26 \\
        D2Prune & 41.02 & 55.99 & 37.47 & 60.59 & 928.95 & 36.94 & 44.93 & 38.54 & 16.60 & 17.68 & 20.58 & 33.47 & 27.90 & 95.61 \\
        EL2N & 43.26 & 54.93 & 35.94 & 54.34 & 898.77 & 39.78 & 18.90 & 34.03 & 15.80 & 16.23 & 18.36 & 34.13 & 28.20 & 88.74 \\
        SemDeDup & 43.09 & 55.72 & 37.31 & \textbf{63.11} & 1115.77 & \textbf{41.32} & \textbf{48.71} & 41.16 & \textbf{17.80} & 19.02 & 19.42 & 33.09 & 27.80 & 99.45 \\
        CLIP & 39.24 & 54.77 & 31.80 & 57.71 & 549.88 & 38.06 & 40.38 & 37.63 & 15.92 & 16.78 & 19.69 & 33.75 & \textbf{29.20} & 89.92 \\
        PRISM & 29.97 & 52.06 & 26.73 & 49.33 & 1106.50 & 35.13 & 16.67 & 29.99 & 17.00 & 19.56 & 19.35 & 35.26 & 27.90 & 86.08 \\
        SCAL & 39.52 & 53.96 & 36.08 & 62.87 & 838.18 & 38.49 & 43.64 & \textbf{43.13} & 15.28 & 17.67 & 19.07 & 34.01 & 27.50 & 93.86 \\
        \rowcolor{lightcyan}
        Data-DPO (Ours) & \textbf{45.19} & \textbf{56.05} & \textbf{40.41} & 61.38 & \textbf{1152.75} & 40.46 & 46.39 & 41.48 & 15.16 & \textbf{20.05} & \textbf{21.25} & \textbf{35.28} & 28.00 & \textbf{100.76} \\
      \bottomrule
      \end{tabular}%
    }
  \end{subtable}

  \vspace{10pt}

  \begin{subtable}{\textwidth}
    \centering
    \caption{10\% data subset.}
    \label{tab:vf_dpo_subset_10}
    \resizebox{\textwidth}{!}{%
      \begin{tabular}{lcccccccccccccc}
      \toprule
      \textbf{Method} & \textbf{GQA} & \textbf{VizWiz} & \textbf{TextVQA} & \textbf{SQA-I} & \textbf{MME} & \textbf{MMB-CN} & \textbf{MMB-EN} & \textbf{AI2D} & \textbf{ChartQA} & \textbf{DocVQA} & \textbf{InfoVQA} & \textbf{MMStar} & \textbf{OCRBench} & \textbf{ARP} \\
      \midrule
        Full Data & 47.30 & 55.88 & 35.83 & 61.43 & 1270.40 & 50.26 & 55.67 & 52.14 & 15.68 & 15.80 & 15.53 & 35.32 & 26.20 & 100.00 \\
        \midrule
        Random & 43.31 & 56.02 & 36.13 & 60.92 & 1152.77 & 39.61 & 31.88 & 44.50 & 16.52 & 17.27 & 18.08 & 34.36 & 27.90 & 95.30 \\
        XMAS & 42.73 & 55.82 & 36.55 & \textbf{64.40} & 1008.41 & 45.19 & 54.38 & 43.95 & 16.20 & 18.39 & 18.19 & 34.28 & 28.50 & 99.31 \\
        COINCIDE & 43.33 & \textbf{57.38} & \textbf{38.70} & 60.83 & 1032.00 & 43.56 & 41.24 & 40.16 & 15.84 & 18.96 & 19.25 & 34.27 & 29.00 & 97.93 \\
        D2Prune & 43.47 & 56.40 & 37.81 & 63.01 & 1146.50 & \textbf{47.68} & 51.80 & 43.01 & 16.44 & 17.38 & 18.78 & 34.73 & 29.00 & 100.49 \\
        EL2N & 44.08 & 56.99 & 38.10 & 47.99 & 1088.62 & 39.52 & 31.70 & 29.92 & 15.64 & 18.78 & 18.78 & 34.94 & 28.30 & 92.67 \\
        SemDeDup & 43.93 & 56.39 & 37.20 & 58.65 & 1236.25 & 47.51 & 45.96 & 39.31 & 15.72 & 18.89 & 17.58 & 36.00 & 28.50 & 98.97 \\
        CLIP & 41.02 & 53.63 & 38.11 & 59.84 & 725.09 & 34.28 & 30.50 & 35.56 & 16.36 & 20.29 & 22.72 & 34.90 & \textbf{29.10} & 94.13 \\
        PRISM & 35.21 & 48.80 & 37.87 & 61.97 & 751.61 & 38.49 & 44.16 & \textbf{44.88} & 16.44 & \textbf{23.76} & \textbf{23.37} & 35.92 & 28.80 & 98.99 \\
        SCAL & 42.73 & 56.26 & 37.68 & 60.88 & 1101.46 & 44.42 & \textbf{54.81} & 44.49 & 14.84 & 18.29 & 16.71 & 33.56 & 27.80 & 97.94 \\
        \rowcolor{lightcyan}
        Data-DPO (Ours) & \textbf{46.06} & 56.38 & 38.62 & 59.89 & \textbf{1322.53} & 43.47 & 49.14 & 42.97 & \textbf{17.48} & 19.05 & 19.76 & \textbf{36.30} & 28.10 & \textbf{102.63} \\
      \bottomrule
      \end{tabular}%
    }
  \end{subtable}

  \vspace{10pt}

  \begin{subtable}{\textwidth}
    \centering
    \caption{15\% data subset.}
    \label{tab:vf_dpo_subset_15}
    \resizebox{\textwidth}{!}{%
      \begin{tabular}{lcccccccccccccc}
      \toprule
      \textbf{Method} & \textbf{GQA} & \textbf{VizWiz} & \textbf{TextVQA} & \textbf{SQA-I} & \textbf{MME} & \textbf{MMB-CN} & \textbf{MMB-EN} & \textbf{AI2D} & \textbf{ChartQA} & \textbf{DocVQA} & \textbf{InfoVQA} & \textbf{MMStar} & \textbf{OCRBench} & \textbf{ARP} \\
      \midrule
        Full Data & 47.30 & 55.88 & 35.83 & 61.43 & 1270.40 & 50.26 & 55.67 & 52.14 & 15.68 & 15.80 & 15.53 & 35.32 & 26.20 & 100.00 \\
        \midrule
        Random & 43.19 & 56.33 & 35.41 & 59.59 & 1230.04 & 42.44 & 38.06 & 41.06 & 15.52 & 16.26 & 16.26 & 32.27 & 27.60 & 93.82 \\
        XMAS & 44.68 & 56.52 & 37.19 & 61.18 & 1012.67 & 40.72 & 46.99 & 44.62 & 15.84 & 17.83 & 17.88 & 34.95 & 27.50 & 97.13 \\
        COINCIDE & 44.28 & 56.44 & 37.52 & 55.73 & 1221.20 & 45.96 & 39.26 & 37.50 & 17.24 & 19.03 & 18.32 & 35.59 & 28.80 & 98.39 \\
        D2Prune & 43.96 & 57.16 & 38.40 & 61.63 & 1059.64 & 44.67 & 49.48 & 44.66 & 15.80 & 17.62 & 18.60 & 34.13 & 28.30 & 98.94 \\
        EL2N & 45.81 & 57.45 & 38.52 & 53.69 & 1105.20 & 46.22 & 39.18 & 43.39 & 14.68 & 19.06 & 17.35 & 34.56 & 28.40 & 96.87 \\
        SemDeDup & \textbf{50.00} & 51.57 & 37.30 & 59.69 & \textbf{1247.46} & 43.90 & 45.45 & 43.13 & 17.04 & 18.66 & 18.93 & 35.24 & 28.00 & 100.35 \\
        CLIP & 41.33 & 55.10 & 38.44 & \textbf{65.25} & 640.09 & 39.60 & 41.67 & \textbf{48.41} & 16.68 & 20.19 & \textbf{21.96} & \textbf{37.57} & \textbf{29.30} & 99.24 \\
        PRISM & 39.25 & 52.24 & \textbf{38.62} & 61.23 & 770.61 & 41.67 & 41.07 & 43.78 & \textbf{17.96} & \textbf{21.12} & 21.00 & 36.22 & 29.10 & 98.64 \\
        SCAL & 43.15 & \textbf{57.90} & 38.47 & 57.56 & 1062.93 & \textbf{46.91} & 43.99 & 36.40 & 15.96 & 18.82 & 16.69 & 35.53 & 27.70 & 96.65 \\
        \rowcolor{lightcyan}
        Data-DPO (Ours) & 46.65 & 50.19 & 37.98 & 62.96 & 1240.62 & \textbf{46.91} & \textbf{53.26} & 46.11 & 17.52 & 19.68 & 18.35 & 35.65 & 28.20 & \textbf{102.70} \\
      \bottomrule
      \end{tabular}%
    }
  \end{subtable}
\end{table*}

\begin{table*}[htbp]
  \centering
  \caption{Main experimental results on LLaVA-CoT using Llama-3.2-11B-Vision-Instruct as the target model. The best result in each column is highlighted in bold.}
  \label{tab:llavacot_llama32_11b_comparison}

  \begin{subtable}{\textwidth}
    \centering
    \caption{5\% data subset.}
    \label{tab:llavacot_llama32_11b_subset_5}
    \resizebox{\textwidth}{!}{%
      \begin{tabular}{lccccccccccccc}
      \toprule
      \textbf{Method} & \textbf{MATH-Vision} & \textbf{We-Math} & \textbf{LogicVista} & \textbf{DynaMath} & \textbf{MMStar} & \textbf{MME} & \textbf{MMBench-EN} & \textbf{ScienceQA-IMG} & \textbf{AI2D} & \textbf{ChartQA} & \textbf{InfoVQA} & \textbf{OCRBench} & \textbf{ARP} \\
      \midrule
        Full Data & 17.01 & 31.90 & 29.91 & 16.23 & 56.99 & 1549.25 & 77.92 & 95.44 & 76.94 & 85.68 & 62.29 & 73.50 & 100.00 \\
        \midrule
        Random & 14.08 & 27.99 & 26.12 & 16.07 & 52.35 & 1405.82 & 76.37 & 90.18 & 73.83 & \textbf{83.88} & 65.16 & 71.50 & 93.98 \\
        XMAS & 15.69 & 33.33 & 32.14 & \textbf{16.11} & 54.28 & 1301.91 & 76.80 & 90.38 & 73.93 & 83.68 & 64.51 & 71.20 & 97.51 \\
        COINCIDE & \textbf{17.63} & 38.22 & 31.03 & 15.27 & 52.86 & 1325.47 & 75.00 & 91.18 & \textbf{74.81} & 83.84 & 64.76 & 71.50 & 98.98 \\
        D2Prune & 17.40 & 34.14 & 28.35 & 14.37 & 52.22 & 1271.98 & 76.03 & 90.18 & 74.68 & 82.96 & 64.44 & 73.90 & 96.36 \\
        EL2N & 15.86 & 38.97 & 33.26 & 15.53 & 52.78 & 1378.95 & 75.95 & 91.03 & 74.19 & 83.40 & 65.34 & 69.20 & 99.13 \\
        SemDeDup & 15.33 & 38.91 & 28.35 & 15.49 & 53.56 & 1334.17 & 76.72 & 90.43 & 73.02 & 83.80 & 65.17 & 70.90 & 97.45 \\
        CLIP & 17.40 & 38.28 & 25.00 & 15.47 & 54.63 & 1258.75 & 76.80 & \textbf{91.42} & 72.73 & 80.72 & 62.10 & \textbf{75.50} & 96.98 \\
        PRISM & 17.01 & 40.63 & 29.02 & 13.43 & \textbf{54.79} & 1353.37 & 75.09 & 89.94 & 72.60 & 77.28 & 66.10 & 74.30 & 97.75 \\
        SCAL & 16.32 & 24.48 & 26.79 & 11.50 & 52.82 & \textbf{1445.83} & 76.55 & 89.04 & 72.93 & 67.08 & \textbf{67.69} & 74.90 & 91.19 \\
        \rowcolor{lightcyan}
        Data-DPO (Ours) & 16.74 & \textbf{47.64} & \textbf{38.84} & 15.07 & 53.82 & 1223.83 & \textbf{77.23} & 90.63 & 74.16 & 83.84 & 66.26 & 69.20 & \textbf{102.73} \\
      \bottomrule
      \end{tabular}%
    }
  \end{subtable}

  \vspace{10pt}

  \begin{subtable}{\textwidth}
    \centering
    \caption{10\% data subset.}
    \label{tab:llavacot_llama32_11b_subset_10}
    \resizebox{\textwidth}{!}{%
      \begin{tabular}{lccccccccccccc}
      \toprule
      \textbf{Method} & \textbf{MATH-Vision} & \textbf{We-Math} & \textbf{LogicVista} & \textbf{DynaMath} & \textbf{MMStar} & \textbf{MME} & \textbf{MMBench-EN} & \textbf{ScienceQA-IMG} & \textbf{AI2D} & \textbf{ChartQA} & \textbf{InfoVQA} & \textbf{OCRBench} & \textbf{ARP} \\
      \midrule
        Full Data & 17.01 & 31.90 & 29.91 & 16.23 & 56.99 & 1549.25 & 77.92 & 95.44 & 76.94 & 85.68 & 62.29 & 73.50 & 100.00 \\
        \midrule
        Random & 15.89 & 29.20 & 30.80 & 15.19 & 50.67 & 1471.44 & 76.55 & 91.77 & 74.48 & 84.12 & 66.30 & 74.20 & 96.85 \\
        XMAS & 16.97 & 35.63 & 32.59 & 15.73 & 53.52 & 1466.86 & 76.98 & 92.02 & 74.03 & 84.68 & 64.31 & 72.10 & 99.79 \\
        COINCIDE & 14.67 & 31.78 & 31.47 & \textbf{16.19} & 49.38 & 1465.80 & \textbf{77.58} & \textbf{92.07} & \textbf{76.04} & \textbf{84.80} & 65.57 & 74.20 & 97.68 \\
        D2Prune & 16.38 & 31.09 & 30.36 & 15.37 & 52.11 & 1449.64 & 77.23 & 91.08 & 74.81 & 83.76 & 64.68 & 72.20 & 97.21 \\
        EL2N & 16.05 & 35.06 & 31.47 & 15.57 & 52.72 & 1365.12 & 76.89 & 91.67 & 75.19 & 84.28 & 65.30 & 72.70 & 98.38 \\
        SemDeDup & 15.82 & 37.53 & 30.80 & 15.59 & 52.69 & \textbf{1528.53} & 76.37 & 91.57 & 75.00 & 84.44 & 64.18 & 74.00 & 99.54 \\
        CLIP & \textbf{17.83} & 30.17 & 28.57 & 15.25 & 53.22 & 1455.55 & 76.55 & 91.22 & 75.29 & 81.12 & 62.29 & 73.60 & 96.89 \\
        PRISM & 16.74 & 39.43 & 31.92 & 13.69 & \textbf{53.56} & 1426.71 & 76.12 & 90.18 & 73.12 & 81.04 & \textbf{68.66} & 74.70 & 99.40 \\
        SCAL & 13.91 & 32.47 & 26.79 & 13.43 & 50.21 & 1470.95 & 76.46 & 90.68 & 74.22 & 63.96 & 67.81 & \textbf{74.90} & 92.83 \\
        \rowcolor{lightcyan}
        Data-DPO (Ours) & 17.17 & \textbf{47.99} & \textbf{37.50} & 15.05 & 53.23 & 1436.62 & 77.15 & 89.39 & 73.80 & 84.36 & 65.85 & 72.60 & \textbf{103.93} \\
      \bottomrule
      \end{tabular}%
    }
  \end{subtable}

  \vspace{10pt}

  \begin{subtable}{\textwidth}
    \centering
    \caption{15\% data subset.}
    \label{tab:llavacot_llama32_11b_subset_15}
    \resizebox{\textwidth}{!}{%
      \begin{tabular}{lccccccccccccc}
      \toprule
      \textbf{Method} & \textbf{MATH-Vision} & \textbf{We-Math} & \textbf{LogicVista} & \textbf{DynaMath} & \textbf{MMStar} & \textbf{MME} & \textbf{MMBench-EN} & \textbf{ScienceQA-IMG} & \textbf{AI2D} & \textbf{ChartQA} & \textbf{InfoVQA} & \textbf{OCRBench} & \textbf{ARP} \\
      \midrule
        Full Data & 17.01 & 31.90 & 29.91 & 16.23 & 56.99 & 1549.25 & 77.92 & 95.44 & 76.94 & 85.68 & 62.29 & 73.50 & 100.00 \\
        \midrule
        Random & 15.56 & 29.54 & 32.37 & 15.02 & 53.44 & 1530.66 & 77.75 & 91.03 & 74.09 & \textbf{85.44} & 66.42 & 74.80 & 98.08 \\
        XMAS & 17.30 & 32.41 & 32.37 & 15.23 & 52.77 & 1485.83 & 76.80 & \textbf{92.27} & 74.61 & 84.68 & 65.48 & 74.20 & 99.25 \\
        COINCIDE & 17.04 & 37.99 & 31.25 & 15.35 & 51.78 & 1494.57 & 76.72 & 92.07 & 75.36 & 84.16 & 65.44 & 74.20 & 100.23 \\
        D2Prune & 16.74 & 27.30 & 33.71 & \textbf{15.99} & 53.32 & 1477.34 & 76.89 & 90.98 & 74.29 & 84.68 & 64.91 & 74.90 & 98.30 \\
        EL2N & 17.14 & 43.56 & 31.47 & 15.29 & 53.19 & \textbf{1570.67} & 77.75 & 91.03 & \textbf{75.74} & 84.48 & 65.95 & 73.50 & \textbf{102.46} \\
        SemDeDup & 16.78 & 34.89 & 30.58 & 15.79 & 53.32 & 1478.05 & 77.23 & 91.52 & 75.06 & 84.32 & 64.90 & 74.40 & 99.41 \\
        CLIP & \textbf{18.19} & 39.08 & 33.71 & 14.43 & \textbf{55.46} & 1477.28 & 76.55 & 91.47 & 73.45 & 82.12 & 64.22 & \textbf{75.70} & 101.27 \\
        PRISM & 16.18 & 37.01 & 31.70 & 12.10 & 53.14 & 1494.04 & 77.15 & 90.83 & 74.42 & 83.12 & 67.87 & 73.80 & 98.22 \\
        SCAL & 13.39 & 25.98 & 31.47 & 15.47 & 51.14 & 1505.28 & \textbf{78.09} & 91.77 & 74.61 & 68.56 & \textbf{68.94} & 74.00 & 94.36 \\
        \rowcolor{lightcyan}
        Data-DPO (Ours) & 13.19 & \textbf{46.61} & \textbf{34.38} & 15.03 & 52.83 & 1500.34 & 76.63 & 90.48 & 75.52 & 85.00 & 66.82 & 71.40 & 101.31 \\
      \bottomrule
      \end{tabular}%
    }
  \end{subtable}
\end{table*}

\begin{table*}[htbp]
  \centering
  \caption{Ablation results of proxy models in selection on Vision-Flan. We compare activation probing on the original LLaVA-v1.5-7B target model with probing on a checkpoint trained on randomly sampled 5\% Vision-Flan data under 5\%, 10\%, and 15\% data budgets. The best result in each column is highlighted in bold.}
  \label{tab:proxy_model_5ckpt_ablation}

  \begin{subtable}{\textwidth}
    \centering
    \caption{5\% data subset.}
    \label{tab:proxy_model_5ckpt_ablation_subset_5}
    \resizebox{\textwidth}{!}{%
      \begin{tabular}{lcccccccccccccc}
      \toprule
      \textbf{Method} & \textbf{GQA} & \textbf{VizWiz} & \textbf{TextVQA} & \textbf{SQA-I} & \textbf{MME} & \textbf{MMB-CN} & \textbf{MMB-EN} & \textbf{AI2D} & \textbf{ChartQA} & \textbf{DocVQA} & \textbf{InfoVQA} & \textbf{MMStar} & \textbf{OCRBench} & \textbf{ARP} \\
      \midrule
        Full Data & 47.30 & 55.88 & 35.83 & 61.43 & 1270.40 & 50.26 & 55.67 & 52.14 & 15.68 & 15.80 & 15.53 & 35.32 & 26.20 & 100.00 \\
        \midrule
        Proxy & 42.88 & 55.74 & 40.00 & 51.56 & 1002.23 & 36.08 & 21.31 & 33.61 & \textbf{16.08} & \textbf{20.76} & 20.40 & 34.12 & \textbf{28.80} & 93.17 \\
        \rowcolor{lightcyan}
        Target & \textbf{45.19} & \textbf{56.05} & \textbf{40.41} & \textbf{61.38} & \textbf{1152.75} & \textbf{40.46} & \textbf{46.39} & \textbf{41.48} & 15.16 & 20.05 & \textbf{21.25} & \textbf{35.28} & 28.00 & \textbf{100.76} \\
      \bottomrule
      \end{tabular}%
    }
  \end{subtable}

  \vspace{10pt}

  \begin{subtable}{\textwidth}
    \centering
    \caption{10\% data subset.}
    \label{tab:proxy_model_5ckpt_ablation_subset_10}
    \resizebox{\textwidth}{!}{%
      \begin{tabular}{lcccccccccccccc}
      \toprule
      \textbf{Method} & \textbf{GQA} & \textbf{VizWiz} & \textbf{TextVQA} & \textbf{SQA-I} & \textbf{MME} & \textbf{MMB-CN} & \textbf{MMB-EN} & \textbf{AI2D} & \textbf{ChartQA} & \textbf{DocVQA} & \textbf{InfoVQA} & \textbf{MMStar} & \textbf{OCRBench} & \textbf{ARP} \\
      \midrule
        Full Data & 47.30 & 55.88 & 35.83 & 61.43 & 1270.40 & 50.26 & 55.67 & 52.14 & 15.68 & 15.80 & 15.53 & 35.32 & 26.20 & 100.00 \\
        \midrule
        Proxy & 45.25 & 55.32 & \textbf{39.87} & 52.60 & 1057.25 & 37.71 & 37.46 & 38.86 & 16.68 & \textbf{20.91} & 19.74 & 36.01 & \textbf{29.20} & 97.76 \\
        \rowcolor{lightcyan}
        Target & \textbf{46.06} & \textbf{56.38} & 38.62 & \textbf{59.89} & \textbf{1322.53} & \textbf{43.47} & \textbf{49.14} & \textbf{42.97} & \textbf{17.48} & 19.05 & \textbf{19.76} & \textbf{36.30} & 28.10 & \textbf{102.63} \\
      \bottomrule
      \end{tabular}%
    }
  \end{subtable}

  \vspace{10pt}

  \begin{subtable}{\textwidth}
    \centering
    \caption{15\% data subset.}
    \label{tab:proxy_model_5ckpt_ablation_subset_15}
    \resizebox{\textwidth}{!}{%
      \begin{tabular}{lcccccccccccccc}
      \toprule
      \textbf{Method} & \textbf{GQA} & \textbf{VizWiz} & \textbf{TextVQA} & \textbf{SQA-I} & \textbf{MME} & \textbf{MMB-CN} & \textbf{MMB-EN} & \textbf{AI2D} & \textbf{ChartQA} & \textbf{DocVQA} & \textbf{InfoVQA} & \textbf{MMStar} & \textbf{OCRBench} & \textbf{ARP} \\
      \midrule
        Full Data & 47.30 & 55.88 & 35.83 & 61.43 & 1270.40 & 50.26 & 55.67 & 52.14 & 15.68 & 15.80 & 15.53 & 35.32 & 26.20 & 100.00 \\
        \midrule
        Proxy & 46.16 & \textbf{56.53} & \textbf{39.47} & 60.49 & 1185.56 & 41.58 & 43.30 & \textbf{46.28} & 17.00 & \textbf{20.50} & \textbf{19.17} & 35.49 & \textbf{28.90} & 101.72 \\
        \rowcolor{lightcyan}
        Target & \textbf{46.65} & 50.19 & 37.98 & \textbf{62.96} & \textbf{1240.62} & \textbf{46.91} & \textbf{53.26} & 46.11 & \textbf{17.52} & 19.68 & 18.35 & \textbf{35.65} & 28.20 & \textbf{102.70} \\
      \bottomrule
      \end{tabular}%
    }
  \end{subtable}
\end{table*}

\begin{table*}[htbp]
  \centering
  \caption{Ablation results on the selection reward. We evaluate different combinations of target model preference, quality score, and marginal diversity on the Vision-Flan dataset using LLaVA-v1.5-7B with a 10\% data subset. The best result in each column is highlighted in bold.}
  \label{tab:reward_ablation_vf}

  \begin{subtable}{\textwidth}
    \centering
    \caption{10\% data subset.}
    \label{tab:reward_ablation_vf_subset_10}
    \resizebox{\textwidth}{!}{%
      \begin{tabular}{lcccccccccccccc}
      \toprule
      \textbf{Method} & \textbf{GQA} & \textbf{VizWiz} & \textbf{TextVQA} & \textbf{SQA-I} & \textbf{MME} & \textbf{MMB-CN} & \textbf{MMB-EN} & \textbf{AI2D} & \textbf{ChartQA} & \textbf{DocVQA} & \textbf{InfoVQA} & \textbf{MMStar} & \textbf{OCRBench} & \textbf{ARP} \\
      \midrule
        Full Data & 47.30 & 55.88 & 35.83 & 61.43 & 1270.40 & 50.26 & 55.67 & 52.14 & 15.68 & 15.80 & 15.53 & 35.32 & 26.20 & 100.00 \\
        \midrule
        DPO & 44.55 & 42.98 & 33.18 & 58.51 & 1223.71 & 42.78 & 49.66 & 40.51 & 15.44 & 13.46 & 17.95 & 34.99 & 24.80 & 92.33 \\
        Quality & 44.05 & \textbf{56.48} & 38.46 & 59.35 & 1158.65 & 41.15 & 39.08 & 37.37 & 17.00 & 19.72 & 19.48 & 35.95 & \textbf{29.60} & 98.96 \\
        Diversity & 44.07 & 55.78 & 37.28 & 54.49 & 1166.17 & 38.32 & 25.95 & 38.27 & 16.64 & \textbf{20.23} & 18.24 & 35.74 & 28.80 & 95.12 \\
        DPO + Diversity & 45.22 & 55.20 & 38.01 & \textbf{62.32} & 1189.84 & 39.00 & 47.94 & \textbf{43.65} & 15.76 & 18.33 & 19.56 & 35.29 & 27.50 & 99.26 \\
        DPO + Quality & 44.25 & 38.24 & 36.84 & 59.64 & 1299.21 & 43.55 & \textbf{52.23} & 41.90 & 16.08 & 12.72 & 16.98 & \textbf{36.48} & 24.00 & 93.25 \\
        Diversity + Quality & 45.48 & 53.35 & \textbf{39.47} & 55.92 & 1167.70 & \textbf{44.67} & 42.18 & 43.17 & 16.76 & 20.02 & 19.15 & \textbf{36.48} & 28.90 & 100.21 \\
        \rowcolor{lightcyan}
        All & \textbf{46.06} & 56.38 & 38.62 & 59.89 & \textbf{1322.53} & 43.47 & 49.14 & 42.97 & \textbf{17.48} & 19.05 & \textbf{19.76} & 36.30 & 28.10 & \textbf{102.63} \\
      \bottomrule
      \end{tabular}%
    }
  \end{subtable}
\end{table*}

\begin{table*}[htbp]
  \centering
  \caption{Robustness results across target models on Vision-Flan using LLaVA-v1.5-13B as the target model. The best result in each column is highlighted in bold.}
  \label{tab:robustness_vf_llava13b}

  \begin{subtable}{\textwidth}
    \centering
    \caption{5\% data subset.}
    \label{tab:robustness_vf_llava13b_subset_5}
    \resizebox{\textwidth}{!}{%
      \begin{tabular}{lcccccccccccccc}
      \toprule
      \textbf{Method} & \textbf{GQA} & \textbf{VizWiz} & \textbf{TextVQA} & \textbf{SQA-I} & \textbf{MME} & \textbf{MMB-CN} & \textbf{MMB-EN} & \textbf{AI2D} & \textbf{ChartQA} & \textbf{DocVQA} & \textbf{InfoVQA} & \textbf{MMStar} & \textbf{OCRBench} & \textbf{ARP} \\
      \midrule
        Full Data & 49.02 & 55.92 & 35.82 & 70.60 & 1318.89 & 56.44 & 60.74 & 57.67 & 19.44 & 20.03 & 19.87 & 35.65 & 29.50 & 100.00 \\
        \midrule
        Random & 43.02 & 56.82 & 34.35 & 63.46 & 1178.91 & 36.25 & 41.23 & 45.66 & 14.88 & 14.87 & 18.17 & \textbf{32.17} & 29.00 & 85.12 \\
        \rowcolor{lightcyan}
        Data-DPO & \textbf{43.46} & \textbf{57.59} & \textbf{39.03} & \textbf{65.29} & \textbf{1276.61} & \textbf{43.99} & \textbf{49.05} & \textbf{46.96} & \textbf{16.52} & \textbf{20.26} & \textbf{22.81} & 30.97 & \textbf{29.80} & \textbf{93.76} \\
      \bottomrule
      \end{tabular}%
    }
  \end{subtable}

  \vspace{10pt}

  \begin{subtable}{\textwidth}
    \centering
    \caption{10\% data subset.}
    \label{tab:robustness_vf_llava13b_subset_10}
    \resizebox{\textwidth}{!}{%
      \begin{tabular}{lcccccccccccccc}
      \toprule
      \textbf{Method} & \textbf{GQA} & \textbf{VizWiz} & \textbf{TextVQA} & \textbf{SQA-I} & \textbf{MME} & \textbf{MMB-CN} & \textbf{MMB-EN} & \textbf{AI2D} & \textbf{ChartQA} & \textbf{DocVQA} & \textbf{InfoVQA} & \textbf{MMStar} & \textbf{OCRBench} & \textbf{ARP} \\
      \midrule
        Full Data & 49.02 & 55.92 & 35.82 & 70.60 & 1318.89 & 56.44 & 60.74 & 57.67 & 19.44 & 20.03 & 19.87 & 35.65 & 29.50 & 100.00 \\
        \midrule
        Random & 44.98 & \textbf{58.13} & 36.43 & \textbf{68.26} & 1176.93 & \textbf{47.60} & \textbf{55.56} & 49.63 & 16.56 & 16.97 & 19.08 & 31.35 & \textbf{29.60} & 92.26 \\
        \rowcolor{lightcyan}
        Data-DPO & \textbf{45.22} & 51.10 & \textbf{38.26} & 66.29 & \textbf{1314.84} & 46.22 & 52.66 & \textbf{52.04} & \textbf{16.80} & \textbf{19.73} & \textbf{21.82} & \textbf{32.42} & \textbf{29.60} & \textbf{94.53} \\
      \bottomrule
      \end{tabular}%
    }
  \end{subtable}

  \vspace{10pt}

  \begin{subtable}{\textwidth}
    \centering
    \caption{15\% data subset.}
    \label{tab:robustness_vf_llava13b_subset_15}
    \resizebox{\textwidth}{!}{%
      \begin{tabular}{lcccccccccccccc}
      \toprule
      \textbf{Method} & \textbf{GQA} & \textbf{VizWiz} & \textbf{TextVQA} & \textbf{SQA-I} & \textbf{MME} & \textbf{MMB-CN} & \textbf{MMB-EN} & \textbf{AI2D} & \textbf{ChartQA} & \textbf{DocVQA} & \textbf{InfoVQA} & \textbf{MMStar} & \textbf{OCRBench} & \textbf{ARP} \\
      \midrule
        Full Data & 49.02 & 55.92 & 35.82 & 70.60 & 1318.89 & 56.44 & 60.74 & 57.67 & 19.44 & 20.03 & 19.87 & 35.65 & 29.50 & 100.00 \\
        \midrule
        Random & 45.25 & \textbf{56.96} & \textbf{38.03} & 65.05 & 1176.15 & 42.70 & \textbf{46.99} & 48.32 & 16.60 & 18.38 & 18.23 & 34.86 & 29.40 & 91.14 \\
        \rowcolor{lightcyan}
        Data-DPO & \textbf{47.75} & 56.18 & 37.70 & \textbf{67.63} & \textbf{1293.21} & \textbf{43.04} & 41.41 & \textbf{49.74} & \textbf{17.92} & \textbf{19.46} & \textbf{21.66} & \textbf{35.30} & \textbf{30.10} & \textbf{94.39} \\
      \bottomrule
      \end{tabular}%
    }
  \end{subtable}
\end{table*}

\begin{table*}[htbp]
  \centering
  \caption{Robustness results across target models on LLaVA-CoT using Qwen2VL-2B-Instruct as the target model. The best result in each column is highlighted in bold.}
  \label{tab:robustness_cot_qwen2vl2b}

  \begin{subtable}{\textwidth}
    \centering
    \caption{5\% data subset.}
    \label{tab:robustness_cot_qwen2vl2b_subset_5}
    \resizebox{\textwidth}{!}{%
      \begin{tabular}{lccccccccc}
      \toprule
      \textbf{Method} & \textbf{MATH-Vision} & \textbf{MME} & \textbf{MMB-EN} & \textbf{SQA-I} & \textbf{AI2D} & \textbf{ChartQA} & \textbf{InfoVQA} & \textbf{OCRBench} & \textbf{ARP} \\
      \midrule
        Full Data & 9.05 & 1350.72 & 64.86 & 77.89 & 63.18 & 73.44 & 49.67 & 71.00 & 100.00 \\
        \midrule
        Random & 9.74 & \textbf{1233.91} & 58.08 & 71.19 & 59.29 & \textbf{69.00} & 45.24 & \textbf{67.90} & 94.30 \\
        \rowcolor{lightcyan}
        Data-DPO & \textbf{15.10} & 955.55 & \textbf{60.48} & \textbf{72.88} & \textbf{60.23} & 68.84 & \textbf{46.09} & 67.30 & \textbf{100.13} \\
      \bottomrule
      \end{tabular}%
    }
  \end{subtable}

  \vspace{10pt}

  \begin{subtable}{\textwidth}
    \centering
    \caption{10\% data subset.}
    \label{tab:robustness_cot_qwen2vl2b_subset_10}
    \resizebox{\textwidth}{!}{%
      \begin{tabular}{lccccccccc}
      \toprule
      \textbf{Method} & \textbf{MATH-Vision} & \textbf{MME} & \textbf{MMB-EN} & \textbf{SQA-I} & \textbf{AI2D} & \textbf{ChartQA} & \textbf{InfoVQA} & \textbf{OCRBench} & \textbf{ARP} \\
      \midrule
        Full Data & 9.05 & 1350.72 & 64.86 & 77.89 & 63.18 & 73.44 & 49.67 & 71.00 & 100.00 \\
        \midrule
        Random & 9.80 & \textbf{1216.21} & \textbf{61.08} & 69.26 & 59.52 & 71.04 & 45.98 & \textbf{68.90} & 95.25 \\
        \rowcolor{lightcyan}
        Data-DPO & \textbf{12.14} & 1039.20 & 60.74 & \textbf{73.18} & \textbf{59.68} & \textbf{71.40} & \textbf{47.10} & 67.30 & \textbf{97.50} \\
      \bottomrule
      \end{tabular}%
    }
  \end{subtable}

  \vspace{10pt}

  \begin{subtable}{\textwidth}
    \centering
    \caption{15\% data subset.}
    \label{tab:robustness_cot_qwen2vl2b_subset_15}
    \resizebox{\textwidth}{!}{%
      \begin{tabular}{lccccccccc}
      \toprule
      \textbf{Method} & \textbf{MATH-Vision} & \textbf{MME} & \textbf{MMB-EN} & \textbf{SQA-I} & \textbf{AI2D} & \textbf{ChartQA} & \textbf{InfoVQA} & \textbf{OCRBench} & \textbf{ARP} \\
      \midrule
        Full Data & 9.05 & 1350.72 & 64.86 & 77.89 & 63.18 & 73.44 & 49.67 & 71.00 & 100.00 \\
        \midrule
        Random & 8.68 & \textbf{1282.13} & 59.97 & 71.99 & 59.49 & 71.52 & 46.41 & \textbf{69.20} & 94.77 \\
        \rowcolor{lightcyan}
        Data-DPO & \textbf{11.91} & 1135.57 & \textbf{61.00} & \textbf{72.88} & \textbf{59.62} & \textbf{71.68} & \textbf{47.20} & 69.00 & \textbf{98.43} \\
      \bottomrule
      \end{tabular}%
    }
  \end{subtable}
\end{table*}

\begin{table*}[htbp]
  \centering
  \caption{Robustness to quality scoring models on Vision-Flan. We compare the default Qwen3-VL-4B-Instruct quality scoring model with LLaVA-OneVision-1.5-4B-Instruct under 5\%, 10\%, and 15\% data budgets. The best result in each column is highlighted in bold.}
  \label{tab:quality_signal_vf}

  \begin{subtable}{\textwidth}
    \centering
    \caption{5\% data subset.}
    \label{tab:quality_signal_vf_subset_5}
    \resizebox{\textwidth}{!}{%
      \begin{tabular}{lcccccccccccccc}
      \toprule
      \textbf{Method} & \textbf{GQA} & \textbf{VizWiz} & \textbf{TextVQA} & \textbf{SQA-I} & \textbf{MME} & \textbf{MMB-CN} & \textbf{MMB-EN} & \textbf{AI2D} & \textbf{ChartQA} & \textbf{DocVQA} & \textbf{InfoVQA} & \textbf{MMStar} & \textbf{OCRBench} & \textbf{ARP} \\
      \midrule
        Full Data & 47.30 & 55.88 & 35.83 & 61.43 & 1270.40 & 50.26 & 55.67 & 52.14 & 15.68 & 15.80 & 15.53 & 35.32 & 26.20 & 100.00 \\
        \midrule
        Random & 42.93 & 55.79 & 36.54 & 60.34 & 1098.02 & 36.40 & 26.19 & 37.82 & \textbf{15.96} & 16.46 & 19.99 & 34.77 & 27.80 & 92.96 \\
        Qwen3-VL-4B & \textbf{45.19} & 56.05 & \textbf{40.41} & 61.38 & \textbf{1152.75} & \textbf{40.46} & 46.39 & 41.48 & 15.16 & \textbf{20.05} & 21.25 & \textbf{35.28} & \textbf{28.00} & 100.76 \\
        LLaVA-OV-4B & 44.24 & \textbf{57.67} & 39.66 & \textbf{64.70} & 1096.94 & 39.35 & \textbf{49.48} & \textbf{44.17} & 15.56 & 19.09 & \textbf{21.61} & 34.95 & \textbf{28.00} & \textbf{101.23} \\
      \bottomrule
      \end{tabular}%
    }
  \end{subtable}

  \vspace{10pt}

  \begin{subtable}{\textwidth}
    \centering
    \caption{10\% data subset.}
    \label{tab:quality_signal_vf_subset_10}
    \resizebox{\textwidth}{!}{%
      \begin{tabular}{lcccccccccccccc}
      \toprule
      \textbf{Method} & \textbf{GQA} & \textbf{VizWiz} & \textbf{TextVQA} & \textbf{SQA-I} & \textbf{MME} & \textbf{MMB-CN} & \textbf{MMB-EN} & \textbf{AI2D} & \textbf{ChartQA} & \textbf{DocVQA} & \textbf{InfoVQA} & \textbf{MMStar} & \textbf{OCRBench} & \textbf{ARP} \\
      \midrule
        Full Data & 47.30 & 55.88 & 35.83 & 61.43 & 1270.40 & 50.26 & 55.67 & 52.14 & 15.68 & 15.80 & 15.53 & 35.32 & 26.20 & 100.00 \\
        \midrule
        Random & 43.31 & 56.02 & 36.13 & \textbf{60.92} & 1152.77 & 39.61 & 31.88 & \textbf{44.50} & 16.52 & 17.27 & 18.08 & 34.36 & 27.90 & 95.30 \\
        Qwen3-VL-4B & 46.06 & \textbf{56.38} & \textbf{38.62} & 59.89 & \textbf{1322.53} & 43.47 & 49.14 & 42.97 & \textbf{17.48} & \textbf{19.05} & 19.76 & \textbf{36.30} & \textbf{28.10} & \textbf{102.63} \\
        LLaVA-OV-4B & \textbf{46.38} & 54.49 & 38.34 & 60.49 & 1198.31 & \textbf{44.50} & \textbf{50.26} & 43.07 & 16.00 & 18.52 & \textbf{20.10} & 33.53 & 27.90 & 100.53 \\
      \bottomrule
      \end{tabular}%
    }
  \end{subtable}

  \vspace{10pt}

  \begin{subtable}{\textwidth}
    \centering
    \caption{15\% data subset.}
    \label{tab:quality_signal_vf_subset_15}
    \resizebox{\textwidth}{!}{%
      \begin{tabular}{lcccccccccccccc}
      \toprule
      \textbf{Method} & \textbf{GQA} & \textbf{VizWiz} & \textbf{TextVQA} & \textbf{SQA-I} & \textbf{MME} & \textbf{MMB-CN} & \textbf{MMB-EN} & \textbf{AI2D} & \textbf{ChartQA} & \textbf{DocVQA} & \textbf{InfoVQA} & \textbf{MMStar} & \textbf{OCRBench} & \textbf{ARP} \\
      \midrule
        Full Data & 47.30 & 55.88 & 35.83 & 61.43 & 1270.40 & 50.26 & 55.67 & 52.14 & 15.68 & 15.80 & 15.53 & 35.32 & 26.20 & 100.00 \\
        \midrule
        Random & 43.19 & 56.33 & 35.41 & 59.59 & 1230.04 & 42.44 & 38.06 & 41.06 & 15.52 & 16.26 & 16.26 & 32.27 & 27.60 & 93.82 \\
        Qwen3-VL-4B & \textbf{46.65} & 50.19 & 37.98 & \textbf{62.96} & \textbf{1240.62} & \textbf{46.91} & \textbf{53.26} & \textbf{46.11} & \textbf{17.52} & 19.68 & 18.35 & 35.65 & 28.20 & 102.70 \\
        LLaVA-OV-4B & 46.50 & \textbf{57.07} & \textbf{39.18} & 61.68 & 1229.97 & 43.21 & 49.23 & 44.85 & 17.48 & \textbf{20.17} & \textbf{20.62} & \textbf{35.84} & \textbf{28.30} & \textbf{103.76} \\
      \bottomrule
      \end{tabular}%
    }
  \end{subtable}
\end{table*}

\begin{table*}[htbp]
  \centering
  \caption{Robustness to embedding sources on Vision-Flan. We compare different embedding sources for Data-DPO using LLaVA-v1.5-7B with 5\%, 10\%, and 15\% data subsets. The best result in each column is highlighted in bold.}
  \label{tab:embedding_source_robustness_vf}

  \begin{subtable}{\textwidth}
    \centering
    \caption{5\% data subset.}
    \label{tab:embedding_source_robustness_vf_subset_5}
    \resizebox{\textwidth}{!}{%
      \begin{tabular}{lcccccccccccccc}
      \toprule
      \textbf{Method} & \textbf{GQA} & \textbf{VizWiz} & \textbf{TextVQA} & \textbf{SQA-I} & \textbf{MME} & \textbf{MMB-CN} & \textbf{MMB-EN} & \textbf{AI2D} & \textbf{ChartQA} & \textbf{DocVQA} & \textbf{InfoVQA} & \textbf{MMStar} & \textbf{OCRBench} & \textbf{ARP} \\
      \midrule
        Full Data & 47.30 & 55.88 & 35.83 & 61.43 & 1270.40 & 50.26 & 55.67 & 52.14 & 15.68 & 15.80 & 15.53 & 35.32 & 26.20 & 100.00 \\
        \midrule
        Random & 42.93 & 55.79 & 36.54 & 60.34 & 1098.02 & 36.40 & 26.19 & 37.82 & 15.96 & 16.46 & 19.99 & 34.77 & 27.80 & 92.96 \\
        2B & 45.19 & 56.05 & \textbf{40.41} & 61.38 & 1152.75 & 40.46 & 46.39 & 41.48 & 15.16 & \textbf{20.05} & \textbf{21.25} & \textbf{35.28} & \textbf{28.00} & \textbf{100.76} \\
        8B & \textbf{45.25} & \textbf{56.69} & 37.86 & \textbf{61.82} & \textbf{1186.49} & \textbf{41.58} & \textbf{50.52} & \textbf{42.16} & \textbf{17.08} & 17.85 & 19.09 & 34.48 & 27.70 & 99.95 \\
      \bottomrule
      \end{tabular}%
    }
  \end{subtable}

  \vspace{10pt}

  \begin{subtable}{\textwidth}
    \centering
    \caption{10\% data subset.}
    \label{tab:embedding_source_robustness_vf_subset_10}
    \resizebox{\textwidth}{!}{%
      \begin{tabular}{lcccccccccccccc}
      \toprule
      \textbf{Method} & \textbf{GQA} & \textbf{VizWiz} & \textbf{TextVQA} & \textbf{SQA-I} & \textbf{MME} & \textbf{MMB-CN} & \textbf{MMB-EN} & \textbf{AI2D} & \textbf{ChartQA} & \textbf{DocVQA} & \textbf{InfoVQA} & \textbf{MMStar} & \textbf{OCRBench} & \textbf{ARP} \\
      \midrule
        Full Data & 47.30 & 55.88 & 35.83 & 61.43 & 1270.40 & 50.26 & 55.67 & 52.14 & 15.68 & 15.80 & 15.53 & 35.32 & 26.20 & 100.00 \\
        \midrule
        Random & 43.31 & 56.02 & 36.13 & 60.92 & 1152.77 & 39.61 & 31.88 & \textbf{44.50} & 16.52 & 17.27 & 18.08 & 34.36 & 27.90 & 95.30 \\
        2B & \textbf{46.06} & 56.38 & \textbf{38.62} & 59.89 & \textbf{1322.53} & \textbf{43.47} & \textbf{49.14} & 42.97 & \textbf{17.48} & \textbf{19.05} & \textbf{19.76} & 36.30 & \textbf{28.10} & \textbf{102.63} \\
        8B & 44.84 & \textbf{56.70} & 37.19 & \textbf{62.72} & 1320.22 & 42.53 & 48.54 & 44.24 & 16.92 & 16.53 & 18.87 & \textbf{36.34} & 27.50 & 100.36 \\
      \bottomrule
      \end{tabular}%
    }
  \end{subtable}

  \vspace{10pt}

  \begin{subtable}{\textwidth}
    \centering
    \caption{15\% data subset.}
    \label{tab:embedding_source_robustness_vf_subset_15}
    \resizebox{\textwidth}{!}{%
      \begin{tabular}{lcccccccccccccc}
      \toprule
      \textbf{Method} & \textbf{GQA} & \textbf{VizWiz} & \textbf{TextVQA} & \textbf{SQA-I} & \textbf{MME} & \textbf{MMB-CN} & \textbf{MMB-EN} & \textbf{AI2D} & \textbf{ChartQA} & \textbf{DocVQA} & \textbf{InfoVQA} & \textbf{MMStar} & \textbf{OCRBench} & \textbf{ARP} \\
      \midrule
        Full Data & 47.30 & 55.88 & 35.83 & 61.43 & 1270.40 & 50.26 & 55.67 & 52.14 & 15.68 & 15.80 & 15.53 & 35.32 & 26.20 & 100.00 \\
        \midrule
        Random & 43.19 & 56.33 & 35.41 & 59.59 & 1230.04 & 42.44 & 38.06 & 41.06 & 15.52 & 16.26 & 16.26 & 32.27 & 27.60 & 93.82 \\
        2B & 46.65 & 50.19 & \textbf{37.98} & \textbf{62.96} & 1240.62 & \textbf{46.91} & \textbf{53.26} & \textbf{46.11} & 17.52 & \textbf{19.68} & 18.35 & 35.65 & \textbf{28.20} & \textbf{102.70} \\
        8B & \textbf{47.27} & \textbf{56.82} & 37.95 & 49.73 & \textbf{1262.70} & 46.22 & 46.13 & 37.21 & \textbf{18.48} & 18.60 & \textbf{20.07} & \textbf{37.29} & 27.60 & 100.75 \\
      \bottomrule
      \end{tabular}%
    }
  \end{subtable}
\end{table*}

\begin{table*}[htbp]
  \centering
  \caption{Sensitivity to reward model capacity on Vision-Flan. We compare Data-DPO with the default residual MLP reward model and a simpler plain MLP reward model using LLaVA-v1.5-7B under 5\%, 10\%, and 15\% data budgets. The best result in each column is highlighted in bold.}
  \label{tab:reward_model_robustness_vf}

  \begin{subtable}{\textwidth}
    \centering
    \caption{5\% data subset.}
    \label{tab:reward_model_robustness_vf_subset_5}
    \resizebox{\textwidth}{!}{%
      \begin{tabular}{lcccccccccccccc}
      \toprule
      \textbf{Method} & \textbf{GQA} & \textbf{VizWiz} & \textbf{TextVQA} & \textbf{SQA-I} & \textbf{MME} & \textbf{MMB-CN} & \textbf{MMB-EN} & \textbf{AI2D} & \textbf{ChartQA} & \textbf{DocVQA} & \textbf{InfoVQA} & \textbf{MMStar} & \textbf{OCRBench} & \textbf{ARP} \\
      \midrule
        Full Data & 47.30 & 55.88 & 35.83 & 61.43 & 1270.40 & 50.26 & 55.67 & 52.14 & 15.68 & 15.80 & 15.53 & 35.32 & 26.20 & 100.00 \\
        \midrule
        Random & 42.93 & 55.79 & 36.54 & 60.34 & 1098.02 & 36.40 & 26.19 & 37.82 & \textbf{15.96} & 16.46 & 19.99 & 34.77 & 27.80 & 92.96 \\
        Plain & 44.26 & \textbf{57.45} & 38.62 & 53.40 & 1145.31 & \textbf{43.21} & 32.47 & 35.85 & 15.36 & 17.53 & 19.27 & \textbf{37.40} & 27.90 & 95.36 \\
        Residual & \textbf{45.19} & 56.05 & \textbf{40.41} & \textbf{61.38} & \textbf{1152.75} & 40.46 & \textbf{46.39} & \textbf{41.48} & 15.16 & \textbf{20.05} & \textbf{21.25} & 35.28 & \textbf{28.00} & \textbf{100.76} \\
      \bottomrule
      \end{tabular}%
    }
  \end{subtable}

  \vspace{10pt}

  \begin{subtable}{\textwidth}
    \centering
    \caption{10\% data subset.}
    \label{tab:reward_model_robustness_vf_subset_10}
    \resizebox{\textwidth}{!}{%
      \begin{tabular}{lcccccccccccccc}
      \toprule
      \textbf{Method} & \textbf{GQA} & \textbf{VizWiz} & \textbf{TextVQA} & \textbf{SQA-I} & \textbf{MME} & \textbf{MMB-CN} & \textbf{MMB-EN} & \textbf{AI2D} & \textbf{ChartQA} & \textbf{DocVQA} & \textbf{InfoVQA} & \textbf{MMStar} & \textbf{OCRBench} & \textbf{ARP} \\
      \midrule
        Full Data & 47.30 & 55.88 & 35.83 & 61.43 & 1270.40 & 50.26 & 55.67 & 52.14 & 15.68 & 15.80 & 15.53 & 35.32 & 26.20 & 100.00 \\
        \midrule
        Random & 43.31 & 56.02 & 36.13 & \textbf{60.92} & 1152.77 & 39.61 & 31.88 & 44.50 & 16.52 & 17.27 & 18.08 & 34.36 & 27.90 & 95.30 \\
        Plain & \textbf{46.41} & 54.19 & \textbf{38.81} & 60.34 & 1242.31 & 43.04 & \textbf{52.92} & \textbf{44.69} & 17.08 & 18.98 & 18.82 & 34.70 & \textbf{28.40} & 101.75 \\
        Residual & 46.06 & \textbf{56.38} & 38.62 & 59.89 & \textbf{1322.53} & \textbf{43.47} & 49.14 & 42.97 & \textbf{17.48} & \textbf{19.05} & \textbf{19.76} & \textbf{36.30} & 28.10 & \textbf{102.63} \\
      \bottomrule
      \end{tabular}%
    }
  \end{subtable}

  \vspace{10pt}

  \begin{subtable}{\textwidth}
    \centering
    \caption{15\% data subset.}
    \label{tab:reward_model_robustness_vf_subset_15}
    \resizebox{\textwidth}{!}{%
      \begin{tabular}{lcccccccccccccc}
      \toprule
      \textbf{Method} & \textbf{GQA} & \textbf{VizWiz} & \textbf{TextVQA} & \textbf{SQA-I} & \textbf{MME} & \textbf{MMB-CN} & \textbf{MMB-EN} & \textbf{AI2D} & \textbf{ChartQA} & \textbf{DocVQA} & \textbf{InfoVQA} & \textbf{MMStar} & \textbf{OCRBench} & \textbf{ARP} \\
      \midrule
        Full Data & 47.30 & 55.88 & 35.83 & 61.43 & 1270.40 & 50.26 & 55.67 & 52.14 & 15.68 & 15.80 & 15.53 & 35.32 & 26.20 & 100.00 \\
        \midrule
        Random & 43.19 & \textbf{56.33} & 35.41 & 59.59 & 1230.04 & 42.44 & 38.06 & 41.06 & 15.52 & 16.26 & 16.26 & 32.27 & 27.60 & 93.82 \\
        Plain & 46.30 & 54.50 & 37.51 & 60.34 & \textbf{1257.74} & 45.96 & 52.41 & \textbf{47.44} & 16.44 & 18.68 & 18.01 & \textbf{36.38} & 27.50 & 101.61 \\
        Residual & \textbf{46.65} & 50.19 & \textbf{37.98} & \textbf{62.96} & 1240.62 & \textbf{46.91} & \textbf{53.26} & 46.11 & \textbf{17.52} & \textbf{19.68} & \textbf{18.35} & 35.65 & \textbf{28.20} & \textbf{102.70} \\
      \bottomrule
      \end{tabular}%
    }
  \end{subtable}
\end{table*}

\clearpage

\begin{table*}[t]
\centering
\caption{
Time cost comparison of different methods on LLaVA-CoT.}
\label{tab:time_cost}
\begin{tabular}{lc}
\toprule
\textbf{Method} & \textbf{GPU Hours} \\
\midrule
XMAS & 15.99 \\
ScalSelect & 8.82 \\
PRISM & 6.70 \\
COINCIDE & 9.06 \\
D2 PRUNE & 12.51 \\
EL2N & 9.57 \\
SemDeDup & 9.55 \\
Data-DPO (Ours) & 19.00 \\
\bottomrule
\end{tabular}
\end{table*}

\bibliography{aaai2027}


\end{document}